\documentclass[journal,onecolumn]{IEEEtran}

\usepackage{graphicx} 
\usepackage{booktabs}
\usepackage{amsmath}
\usepackage{mathrsfs}
\usepackage{amsfonts}
\usepackage{amssymb}
\usepackage{epstopdf}
\usepackage[caption=false,font=footnotesize]{subfig}
\usepackage{color}
\usepackage{enumitem}
\usepackage{amsthm}
\usepackage{cite}

\newcommand{\RI}[1]{\textcolor{black}{#1}}
\newcommand{\RII}[1]{\textcolor{black}{#1}}
\newcommand{\RIII}[1]{\textcolor{black}{#1}}

\begin{document}

\title{Cross-subject Decoding of Eye Movement Goals from Local Field Potentials}

\author{\IEEEauthorblockN{Marko~Angjelichinoski\IEEEauthorrefmark{1}, John Choi\IEEEauthorrefmark{2}, Taposh Banerjee\IEEEauthorrefmark{3},\\ Bijan Pesaran\IEEEauthorrefmark{2} and Vahid Tarokh\IEEEauthorrefmark{1}}\\
\IEEEauthorblockA{\IEEEauthorrefmark{1}Department of Electrical and Computer Engineering, Duke University}\\
\IEEEauthorblockA{\IEEEauthorrefmark{2}Center for Neural Science, New York University}\\
\IEEEauthorblockA{\IEEEauthorrefmark{3}Department of Electrical and Computer Engineering,\\University of Texas at San Antonio}
}

\maketitle

\begin{abstract}
\emph{Objective.} We consider the cross-subject decoding problem from local field potential (LFP) signals, where training data collected from the prefrontal cortex (PFC) of a source subject is used to decode intended motor actions in a destination subject. \emph{Approach.} We propose a novel supervised transfer learning technique, referred to as \emph{data centering}, which is used to adapt the feature space of the source to the feature space of the destination. The key ingredients of data centering are the transfer functions used to model the deterministic component of the relationship between the source and destination feature spaces. We propose an efficient data-driven estimation approach for linear transfer functions that uses the first and second order moments of the class-conditional distributions. \emph{Main result.} We apply our data centering technique with linear transfer functions for cross-subject decoding of eye movement intentions in an experiment where two macaque monkeys perform memory-guided visual saccades to one of eight target locations. The results show peak cross-subject decoding performance of $80\%$, which marks a substantial improvement over random choice decoder. In addition to this, data centering also outperforms standard sampling-based methods in setups with imbalanced training data. \emph{Significance.} The analyses presented herein demonstrate that the proposed data centering is a viable novel technique for reliable LFP-based cross-subject brain-computer interfacing and neural prostheses.
\end{abstract}

\IEEEpeerreviewmaketitle

\section{Introduction}\label{sec:intro}

Recent advances in neural engineering have shown that local field potential (LFP) signals collected from the prefrontal cortex (PFC) of a subject via chronically implanted microelectrode arrays, are a reliable alternative to spike count recordings for designing robust and resilient brain-computer interfaces (BCIs) \cite{Stavisky:2015fo,Linden:2011ck,Pesaran:2018,Angjelichinoski2019}.
The theory of non-parametric regression has proven to be crucial for the success of LFP-based decoders.
As documented in \cite{Banerjee1,Angjelichinoski2019}, the application of non-parametric regression to LFPs has led to the development of complex spectrum-based feature extraction technique based on the famous Pinsker's theorem.
As opposed to popular feature extraction methods such as the conventional Power Spectrum Density (PSD)-based decoders \cite{Markowitz18412}, or decoders based on the spatial covariance matrices of the trials \cite{CSP_2000,RiemannianGeo_2012} which take into account only the information stored in the LFP signal amplitude, Pinsker's feature extraction methodology naturally incorporates both the amplitude and the phase information inherently present in the LFP signals. As a result, even simple linear classification methods when applied over Pinsker's features exhibit significant performance improvements in decoding intended motor actions \cite{Angjelichinoski2019}.

\RI{One of the most important objectives of neural engineering is to design robust and reliable \emph{cross-subject} BCIs that generalize well over large population of subjects \cite{Jayaram_2016,Lotte_2018,Zanini2018,Azab_2019}}.
Cross-subject BCIs should be trained over limited number of representative subjects performing same or similar tasks; however, they are being expected to perform reliably over unseen representatives from the same population.
There are several reasons for development of cross-subject BCIs which can be summarized as follows \cite{Lotte_2018}. First, cross-subject BCIs can be used for quantifying similarities between parts of the PFC that determine decision-making across different subjects - an important task in developing cross-subject BCIs that can be used across new subjects from the population without extensive training.
Second, in lack of sufficient training data to train a reliable decoder for any given representative of a population, the cross-subject BCIs should enable and facilitate decoding of intended actions even in absence of local training. This can be useful in subjects with impaired or lost motor functions where training data collected from healthy subjects can be used to alleviate the chronic impairment.

At the heart of such BCIs is the cross-subject decoding problem in which training data collected from one representative subject (henceforth referred to as subject X or source) is used to train a reliable and well-performing decoder for another subject (i.e., subject Y or destination) over specific set of actions. 
In comparison with the subject-specific decoding problem, which has been extensively studied in both non-invasive (such as electroencephalogram (EEG)) \emph{and} invasive BCI configurations, the cross-subject problem has been covered to a lesser extent especially in invasive configurations. 
Regardless of the BCI configuration, even in its most basic variant where the decoder is defined over the same set of actions, the cross-subject decoding turns out to be a challenging neurological problem \cite{Jayaram_2016,Lotte_2018}.
The main reason for its general difficulty is the fact that brain signals are highly variable and their structural and statistical properties can vary dramatically even under slight changes of the recording conditions \cite{FLotte_2015,Kwon_deepConv_2019}.
In fact, in our studies we have observed that a decoder trained over data set collected from given subject in general performs close to random choice decoder when applied directly to test data set from different subject, even in the case where the subjects perform the exact same tasks under the same experimental conditions.

\RI{\emph{Review of existing work.} 
Despite the challenges imposed by the non-stationary nature of neuronal activity signals, the cross-subject decoding problem in non-invasive, EEG-based BCIs has been addressed and the progress has been reported.
To this end, methods from the emerging field of \emph{transfer learning} have been extensively applied, albeit to varying degrees of success \cite{Jayaram_2016}.
In its most general, transfer learning refers to a set of data mining and machine learning techniques, procedures and algorithms designed to extrapolate knowledge acquired in a given domain and apply it in different but somewhat related domain \cite{TransferL_survey1,Weiss2016ASO,Jayaram_2016}.
The \emph{a priori} assumption in transfer learning is that there exist inherent connections, correlations and/or similarities between the domains, and the objective of any transfer learning algorithm is to discover these similarity structures and foster reliable transfer of knowledge across the domains \cite{TransferL_survey1}.
It should be also noted that transfer learning is often done in a non-parametric, data-driven manner without relying extensively on detailed statistical models; this has proven to be effective in problems characterized with highly non-stationary temporal and/or spatial behaviour.}

\RI{In the context of cross-subject BCIs, which can be seen as one specific example of transfer learning, several techniques for identifying and estimating structural similarities between neurological data collected in different subjects have emerged; they can be organized into several broader categories which we briefly outline and discuss next.
The \emph{domain adaptation} is by far the most popular transfer learning methodology for cross-subject BCIs.
As the name itself suggests, the main objective in domain adaptation is to find an invariant, subject-independent data/feature space in which the information-bearing patterns that drive the dynamics of the motor responses remain unchanged across subjects.
Technically, this amounts to finding the corresponding transformation(s) through which source and destination data/feature spaces can be mapped onto the invariant space.
Domain adaptation has been extensively applied over feature spaces based on the spatial covariance matrices such as the Common Spatial Patterns (CSP) method \cite{CSP_2000,FARQUHAR20091278}, which is a popular method for feature extraction from multichannel EEG signals, as well as its more sophisticated extensions based on Riemannian geometry \cite{RiemannianGeo_2012,BARACHANT2013172,GAUR2018201,Yger2017}.
In the case of CSP-based domain adaptation, the usual approach is to assume that there exists a unique bank of spatial filters which is invariant across subjects. Notable work along this line has been reported in \cite{Fazli2009} where the authors use a large database of EEG data collected from multiple source subject performing binary motor tasks to estimate the invariant spatial filters which are used to project the test features of destination subjects onto the invariant feature space where training and decoding takes place; these ideas have been further explored and extended in \cite{Kang2014,LotteGuan2011} as well as \cite{Samek2013,Samek2014,Vidaurre2011,Sugiyama:2007:CSA:1314498.1390324,Sun:2016:RFE:3016100.3016186}.
Similar to spatial filtering, transfer learning in the Riemannian manifold of the spatial covariance matrices has been also considered in \cite{Zanini2018} where the authors assume that different subjects/sessions/tasks induce shifts in the trial covariance matrix with respect to a resting state covariance matrix which is further assumed to be common for all subjects. 
Due to the existing body of work on decoding methods based on Riemannian geometry, Riemannian transfer learning of spatial covariance matrices is a highly active area of research in non-invasive EEG-based BCIs \cite{10.3389/fninf.2018.00066,Rodrigues2019,Maman2019}.
In addition to EEG, domain adaptation has also been reported for cross-subject decoding of other non-invasive signal modalities such as functional magnetic resonance imaging (fMRI) as in \cite{HAXBY2011404}.}

\RI{Another popular group of transfer learning methods for cross-subject BCIs is the \emph{ensemble learning} (also known as rule adaptation \cite{Jayaram_2016}).
Different from domain adaptation where one attempts to find an invariant, common feature space where the subject-specific data is postulated to follow subject-invariant distribution, ensemble learning aims to learn an underlying structure in the space of classification rules \cite{NIPS2012_4775,pmlr-v9-alamgir10a,Jayaram_2016,Yalin2019,Azab_2019}.
Compared to domain adaptation, ensemble learning offers increased flexibility.
Specifically, domain adaptation seeks for a common space where a single decoding rule for all subjects is valid. This is indeed a strong assumption since such space can be difficult to find, might require large number of source subjects to estimate or might not exist at all.
Ensemble learning, on the other hand, aims to capture and model the variations that occur across individual subjects with the objective of finding adequate transformation(s) between the optimal decision rules that govern the decoding performance in each individual subject.
One popular ensemble learning approach is to formulate a joint learning problem with single objective function defined across all subjects and determine the optimal decision rule parameters for each individual subject through joint optimization as in \cite{Jayaram_2016,Yalin2019,Azab_2019}.
A closely related concept is meta-learning, which is conversationally known as ``learning to learn''. Meta-learning is primarily applied in deep learning where the objective is to find common model parameters over a set of individual learning problems, from which the optimal, problem-specific parameters can be reached within few steps of gradient descent optimization; meta-learning quickly became one of the most exciting new developments in deep learning (see the recent surveys \cite{Vanschoren2019,Wang2019} for an in-depth literature review and applications in BCIs).
}

\RI{It should be noted that most of the existing literature on cross-subject decoding covers non-invasive BCI configurations that primarily use multichannel EEG signals.
Nevertheless, few recent contributions address electrophysiology-based studies where chronically implanted microelectrode arrays are used to record the activity of hundreds (even thousands) of neurons \cite{Dyer080861,Pandarinath152884,sussillo2016making}.
An early work reported in \cite{sussillo2016making} uses recurrent neural networks to capture the temporal variability of firing patters of neurons recorded across long period of time, fostering successful cross-session decoding.
In similar fashion, further progress in cross-session decoding has been reported in \cite{Pandarinath152884} where latent factor analysis via dynamical systems was applied to model and predict single-trial firing patterns from large population of neurons.
However, the studies cited above apply subject-specific training and apart from the preliminary work reported in \cite{Dyer080861}, we are not familiar with other dedicated work on transfer learning for cross-subject decoding that uses electrophysiology such as LFPs.
}


\emph{Our contributions.}
In this paper, we introduce novel supervised transfer learning method, tailored for cross-subject decoding of intended motor actions from LFP recordings in Pinsker's feature space \cite{Angjelichinoski2019}. We refer to the method as \emph{data centering}; 
its pivotal idea is to adapt the feature space of the source subject where the training data is collected to the feature space of the destination subject where the decoding of intended motor intentions takes place. The adaptation procedure relies on the notion of \emph{transfer functions} - a concept that formally captures the variability of feature spaces across subjects through functional transformations.
We propose a simple, data-driven solution for estimating \emph{linear} transfer functions that uses only the first and the second order moments of the class-conditional distributions in the feature space. 
We apply data centering with linear transfer functions to the experiment and data introduced in \cite{Markowitz18412} where two adult macaque monkeys perform memory-driven visual saccades to one of eight peripheral target lights under the same experimental conditions.
The results show that when data centering with linear transfer functions is applied over Pinsker's feature space, the cross-subject decoder exhibits substantial performance improvement over random choice decoder, peaking at $80\%$.
We also show how to apply data centering for neural decoding with imbalanced data sets; is such setups, data centering is used to fix the between-class imbalance by bringing additional data for the under-represented class(es) from different source with well-represented classes.
Our investigations demonstrate that data centering outperforms standard sampling-based methods for imbalanced data sets, further proving its promising potential for robust cross-subject BCIs and neural prostheses.


\section{Methods}

\subsection{Description of the Experiment}\label{sec:experiment}

\begin{figure}[h]
\centering
\subfloat[Memory-driven visual saccades experiment]{\includegraphics[scale=0.275]{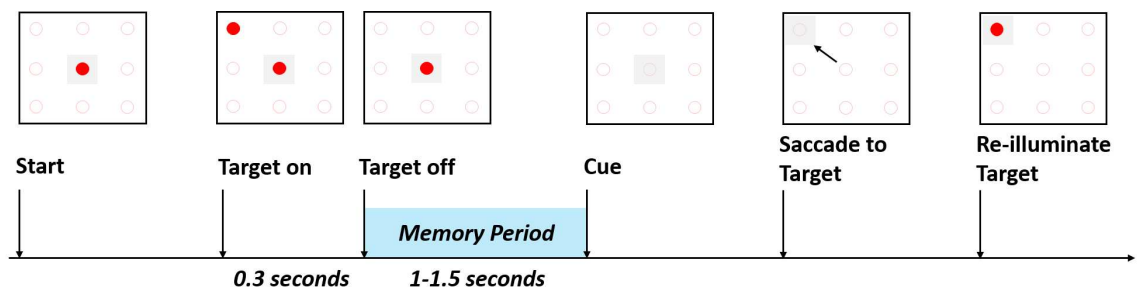}\label{experiment}}
\hfil
\subfloat[Recording chambers and electrode penetration sites]{\includegraphics[scale=0.2]{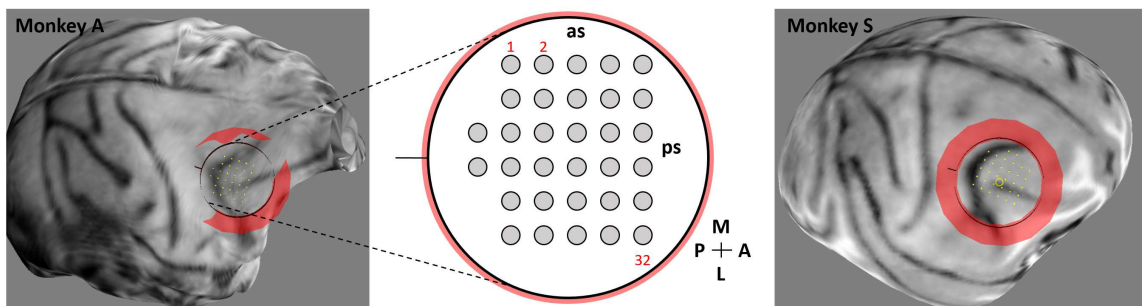}\label{Monkey_brain}}
\caption{\RIII{Experimental setup. (a) Memory-guided visual saccades.  The trial begins when the subject fixates the gaze at the central light displayed on the LCD monitor. A randomly chosen peripheral target light is turned on for the duration of $300$ milliseconds. Turning off the target light marks the beginning of the memory period. After approximately $1$ second, the central light is turned off, giving the subject a cue to saccade to the remembered location of the previously lit target light. (b) Recording chamber locations and electrode (i.e., channel) penetration sites. The chambers were surgically implanted in a craniotomy made over the right prearcuate gyrus of lateral PFC using MRI-guided stereotaxic techniques (Brainsight). The chambers were implanted at (approximately) the same locations in the PFCs of both subjects, covering the same areas of the brain. A microelectrode array consisting of $32$ individually movable channels (20 mm travel, Gray Matter Research) was semichronically implanted in the chamber. Channel numbers are indexed $1-32$ from left to right (Sulcal landmarks: \emph{as}-arcuate sulcus, \emph{ps}-principal sulcus; Compass labels: \emph{A}-anterior, \emph{P}-posterior, \emph{M}-medial, \emph{L}-lateral). The location of the array in Monkey S was rotated by approximately $90$ degrees with respect to Monkey A. The animals were instrumented with a head restraint prosthesis that enables head position fixation and eye movement tracking. See also \cite{Markowitz11084,Markowitz18412} for further details.}}
\label{brains}
\end{figure}

We begin with a brief overview of the experimental setup, schematically depicted in Fig.~\ref{brains} (see also \cite{Markowitz18412, Markowitz11084, Angjelichinoski2019} for further details).
All experimental procedures were approved by the NYU University Animal Welfare Committee (UAWC).

Two adult male macaque monkeys (\emph{M. mulatta}), referred to herein as Monkey A and Monkey S, were trained to perform a memory-guided saccade task which is schematically depicted in Fig.~\ref{experiment}. 
The setup consists of an LCD monitor on which targets are presented to the animal. The monkey initiates a trial by fixating a central target. Once the monkey maintains fixation for a baseline period, one of 8 possible peripheral targets (drawn from the corners and edge midpoints of a square centered on the fixation target) is flashed on the screen for 300 ms. For a random-duration memory period ($1$ - $1.5$ s), the monkey must maintain fixation on the central target until it is turned off. This event cues the monkey to make a saccade to the remembered location of the peripheral target. If the monkey holds his gaze within a square window of the true target location, the trial is completed successfully and the monkey receives a reward; regardless of success or failure, the true target location is reilluminated after the trial as feedback.
In our analyses, only LFP segments during the memory periods of successful trials were used in the decoding experiments.
This epoch of the trial is especially interesting since it reflects information storage and generation of the resulting motor response. 

\RIII{The neuronal activity during the memory period was recorded using a microelectrode array with 32 individually movable electrodes (also referred to as channels), semichronically implanted in a recording chamber in the PFC, see Fig.~\ref{Monkey_brain}. Note that the recording chambers were placed at the same locations of the PFC in both subjects.}
Signals were recorded at $30$ kHz and downsampled to $1$ kHz.
{The initial position of each electrode at the beginning of the experiment was receded $1$ millimeter within the drive; after penetrating the dura and the pia, action potentials were first recorded at a mean depth (from the initial position) of $2.23$ and $3.04$ millimetres for Monkey A and Monkey S, respectively. 
As the experiment progressed, the positions of individual electrodes were gradually advanced deeper; the mean depth increase step was $34$ and $100$ microns for Monkey A and Monkey S, respectively.
For a fixed electrode depth configuration, multiple trials were performed, which is detailed in Section~\ref{sec:DataPrepar}.
Henceforth, a fixed configuration of electrode positions over which trials are collected is referred to as \emph{electrode depth configuration (EDC)}.
Each EDC is uniquely described by a $32$-dimensional real vector; each entry in the vector contains the depth position of an individual electrode (in millimeters) with respect to its initial position. 
}


\subsection{Pinsker's Theorem for Extracting Features from LFPs}\label{sec:pinsker}

We give a brief, informal digest of the feature extraction technique based on the famous Pinsker's theorem; for more rigorous theoretical treatment of the theorem and related concepts such as Gaussian sequence models and minimax optimality, we advise the interested reader to refer to \cite{Johnstone2012GaussianE, Tsybakov:2008:INE:1522486} as well as \cite{Banerjee1,Angjelichinoski2019} where Pinsker's theorem was first applied for extracting features from noisy LFP signals.

\subsubsection{Nonparametric Regression Framework for LFP signals}\label{sec:nonparamLFP}

Let $\tilde{f}_t,t=0,\hdots,T-1$ denote the discrete-time LFP signal from an arbitrary channel (i.e., electrode) sampled with frequency $\nu_S=1$ kHz during the memory period of the memory-guided visual saccade experiment, see Section~\ref{sec:experiment} and Fig.~\ref{experiment}.
We use arguments from the theory of nonparameteric regression to construct a technique for extracting the relevant features from the noisy LFP data \cite{Banerjee1,Angjelichinoski2019}. 

First, we assume that each LFP signal consists of two components: (1) useful, information-carrying and unknown signal waveform, represented with the function $f$, and (2) random noise component $\sigma \omega$ which is modelled as \RIII{independent and identically distributed (i.i.d.)} Gaussian noise. These two components add up in the following model:
\begin{align}\label{eq:lfp_model}
    \tilde{f}_t = f_t + \sigma \omega_t,\quad \omega_t\sim\mathcal{N}(0,1),\quad t=0,\hdots,T-1,
\end{align}
where $f_t = f(t\nu_S)$ and $\omega_t$ are the corresponding discrete versions of $f$ and $\omega$ respectively.
The unknown function $f$ stores the relevant information from which the motor intentions of the subject can be decoded. As such, each different action, i.e., in our case, each different eye movement direction will yield different representation in the function space.
Moreover, as noted in \cite{Angjelichinoski2019}, the signal $f$ will be also different in repeated tasks due to variety of neurological and practical reasons. Hence, it is more accurate to say that each specific motor action when performed repeatedly forms a class of functions in the function space; in such case the neural decoding problem reduces to conventional, multiple-class composite hypothesis testing where the goal is to design a decoder from LFP data that reliably identifies the function class of incoming LFP signal.

Next, the decoder, being a function of random and finite LFP sequences, should be consistent. One way to ensure consistency, is to take the worst case probability of miss-classification to zero as $T\rightarrow\infty$. This motivates the use of minimax-optimal estimators for $f$ from $\tilde{f}_t,t=0,\hdots,T-1$; it can be shown \cite{Banerjee1,Angjelichinoski2019} that using minimax-optimal function estimators leads to a consistent decoder as long as the function classes in the function space are well separated.

\subsubsection{Gaussian Sequences and Pinsker's Theorem}\label{sec:GaussianSeqPinsker}

Function estimators are objects of infinite dimension and therefore hard to handle and use in practice for training.
The Gaussian sequence modeling framework provides an alternative approach for obtaining finite-dimensional, asymptotically minimax-optimal estimators of arbitrary functions as long as these functions satisfy some mild smoothness criteria \cite{Johnstone2012GaussianE,Tsybakov:2008:INE:1522486}. The key idea is to represent the time-domain LFP data \eqref{eq:lfp_model} as a sequence of independent Gaussian random variables with different means; this is achieved by transforming the original, noisy LFP data using common orthogonal transformation such as the Fourier basis.
As a result, we obtain the equivalent, sequence representation of \eqref{eq:lfp_model}:
\begin{align}\label{eq:Gaussian_sequence}
    \tilde{X}_l = X_l + \frac{\sigma}{\sqrt{T}}\Omega_l,\quad l=1,2,\hdots.
\end{align}
Here, $\tilde{X}_l$, $X_l$ and $\Omega_l$ are projections of the vectors $\tilde{f}=(\tilde{f}_0,\hdots,\tilde{f}_{T-1})$, $f=(f_0,\hdots,f_{T-1})$ and $\omega=(\omega_0,\hdots,\omega_{T-1})$ onto the $l$-th Fourier basis function \cite{Johnstone2012GaussianE,Tsybakov:2008:INE:1522486}.
Now, the problem of finding an estimator for the function $f$ reduces to finding an estimate for the Fourier coefficients $X_l,l=1,2,\hdots$ in the equivalent sequence space.

Pinsker's theorem gives an asymptotically minimax-optimal estimator for the Gaussian sequence model provided that the Fourier coefficients satisfy some predefined criteria \cite{Johnstone2012GaussianE,Tsybakov:2008:INE:1522486}. Specifically, if the Fourier coefficients live in (or on) an ellipsoid, i.e., if the sequence $X_l,l=1,2,\hdots$ satisfies $\sum_l a_l^2X_l^2\leq C$ with $a_1=0$, $a_{2k}=a_{2k+1}=(2k)^{\alpha}$ (or, equivalently, the functions $f$ live in a Sobolev space of order $\alpha$), then the asymptotically minimax-optimal estimator of the coefficients $X_l,l=1,2,\hdots$ is given by the simple linear estimator
\begin{align}\label{eq:pinsker_estimator}
    {X}_l = \underbrace{\left(1 - \frac{a_l}{\mu}\right)_+}_{c_l} \tilde{X}_l,\quad l=1,2,\hdots
\end{align}
The parameters $\mu>0$ and $\alpha$ become design parameters and their values need to be carefully chosen such that the performance of the decoder is optimized.
Inspecting the sequence $c_l,l=1,2,\hdots$, we see that the estimator \eqref{eq:pinsker_estimator} shrinks the observations $y_l$ by an amount $1-\frac{a_l}{\mu}$ when $\frac{a_l}{\mu}<1$; otherwise it sets the observations to $0$.
In light of this, it has also been shown that the simple truncation estimator of the form
\begin{eqnarray}\label{eq:truncation}
X = \textup{diag}(1_{l\leq L})\tilde{X},
\end{eqnarray}
is also asymptotically minimax-optimal \cite{Johnstone2012GaussianE}.
Here, $1_{l\leq L}$ is a vector where the first $L<T$ entries are $1$ and the remaining $0$ and the truncation estimator can be viewed as a special case of Pinsker's estimator \eqref{eq:pinsker_estimator} with $c_l=1$ for $l\leq L$ and $c_l=0$ otherwise; in other words, our finite dimensional representation of the estimate of $f$ is obtained by simply retaining the $L$ dominant components of the complex spectrum of the noisy LFP signal. Note that for an LFP sequence of length $T$ samples, acquired with sampling frequency $\nu_S$, the truncation estimator corresponds to a low-pass filter with cut-off frequency $\frac{(L-1)\cdot \nu_S}{T}$.
The main difference between Pinsker's \eqref{eq:pinsker_estimator} and the truncation estimator \eqref{eq:truncation} arises in their respective rates of convergence; namely, Pinsker's estimator converges fastest to the true LFP waveform as $T\rightarrow\infty$ among all minimax-optimal estimators \cite{Johnstone2012GaussianE}. In practice, this is a rather subtle difference and one should not expect significant deviation in the decoding performance between \eqref{eq:pinsker_estimator} and \eqref{eq:truncation}, which our experiments have confirmed \cite{Angjelichinoski2019}. The truncation estimator is also simpler to implement than Pinsker's estimator since it introduces only a single design parameter, namely the number of retained Fourier coefficients $L$. 

Pinsker's estimator has proven to be very useful in limited data scenarios, especially when the number of trials is not more than an order of magnitude larger than the respective dimension of the problem; such data sets frequently arise in electrophysiology-based neuroscientific studies where the cost of running experiments and collecting data is high.
However, in situations where the amount of available data is relatively large, \RI{alternative feature extracting methodologies can be also considered, including non-linear ones such as recurrent neural networks and autoencoders \cite{sussillo2016making, Pandarinath152884, Goodfellow:2016:DL:3086952, Bishop:2006:PRM:1162264}.}

\subsubsection{The Statistical Properties of LFP signals}

\begin{figure*}[h]
\centering
\subfloat{\includegraphics[scale=0.275]{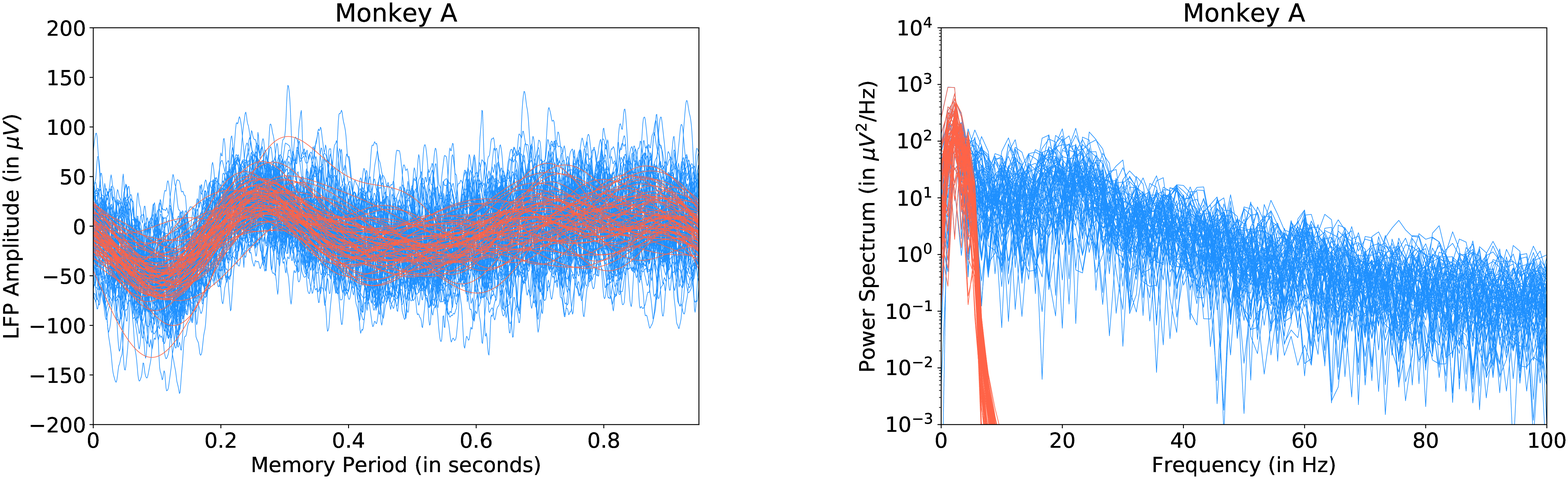}\label{MonkeyA_SignalsPSD}}
\hfil
\subfloat{\includegraphics[scale=0.275]{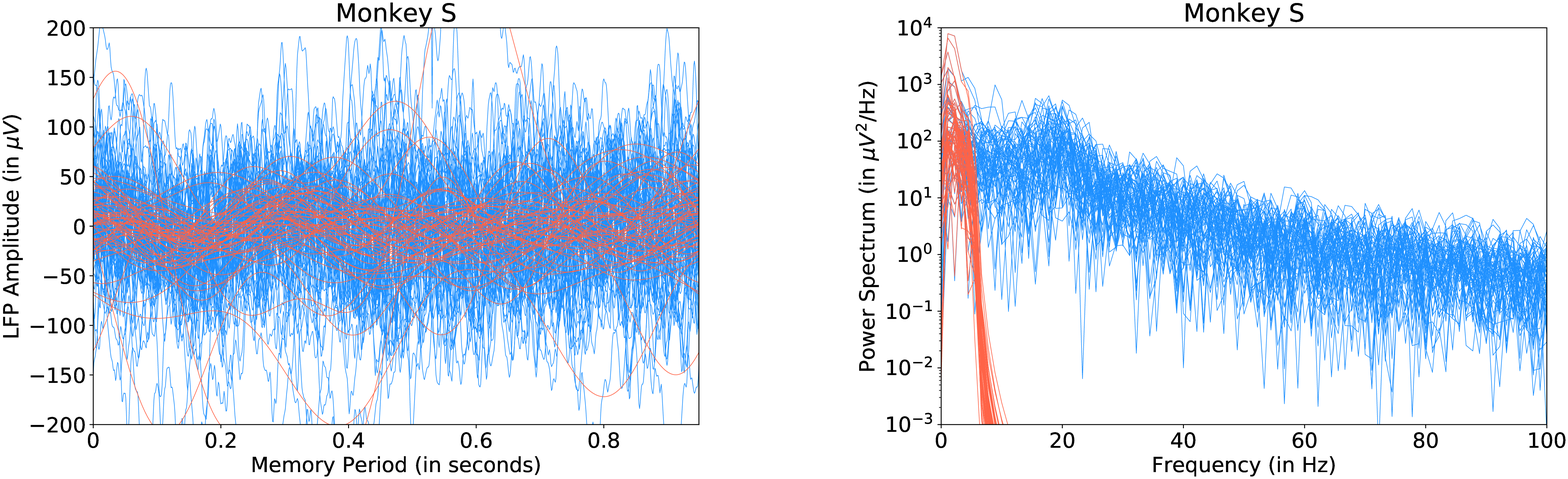}\label{MonkeyS_SignalsPSD}}
\caption{\RI{The LFP signals and their spectral densities. The channel with index $10$ (see Fig.~\ref{Monkey_brain} for details on the channel indexing) and eye movement direction ``$\leftarrow$'' (see Fig.~\ref{experiment} for details on eye movement directions) were arbitrarily selected; same conclusions apply to any other channel and/or other movement direction. Left-hand plots: blue lines correspond to original LFP sequences with $T=950$, red lines correspond to low-pass filtered waveforms using \eqref{eq:truncation} with $L=5$. Right-hand plots: Spectral densities of the original (blue lines) and filtered LFPs (red lines). The PSD was estimated using Welch's method \cite{Welch} with Hann window, segment size of $900$ and overlap of $50\%$.}}
\label{LFPSignalsAndTheirPSDs}
\end{figure*}

The successful application of the non-parametric regression framework to LFP data depends crucially on the validity of the model \eqref{eq:lfp_model} as well the extent to which the LFP signals satisfy the smoothness requirements in Pinsker's theorem. Before moving on to our main contributions, it is particularly important to confirm that the LFP data adheres to the requirements of the non-parametric framework outlined in Sections~\ref{sec:nonparamLFP} and \ref{sec:GaussianSeqPinsker} and in order to do this, we analyse the statistical properties of the LFP signals.

\begin{figure}
\centering
\subfloat{\includegraphics[scale=0.5]{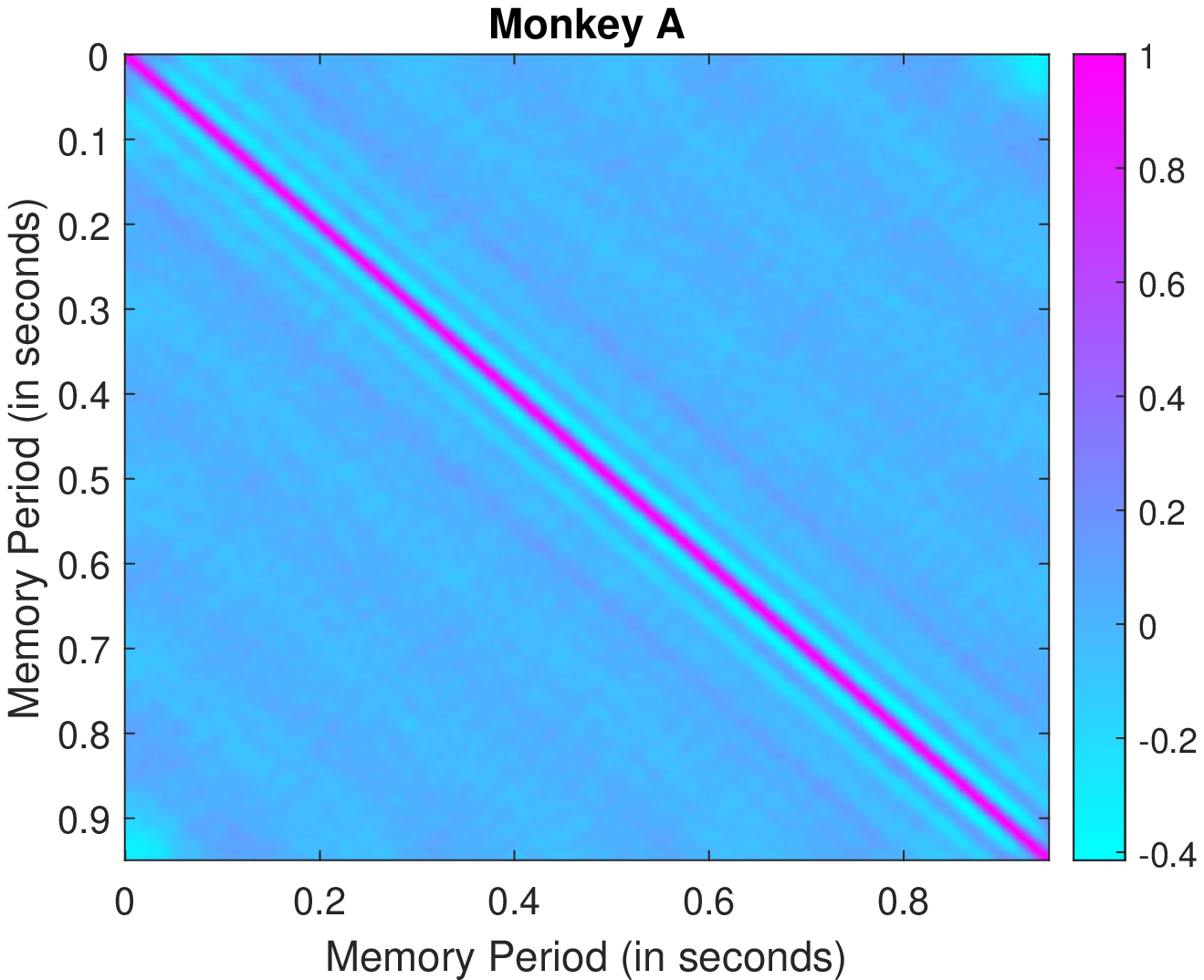}\label{MonkeyA_corrMat}}
\hfil
\subfloat{\includegraphics[scale=0.5]{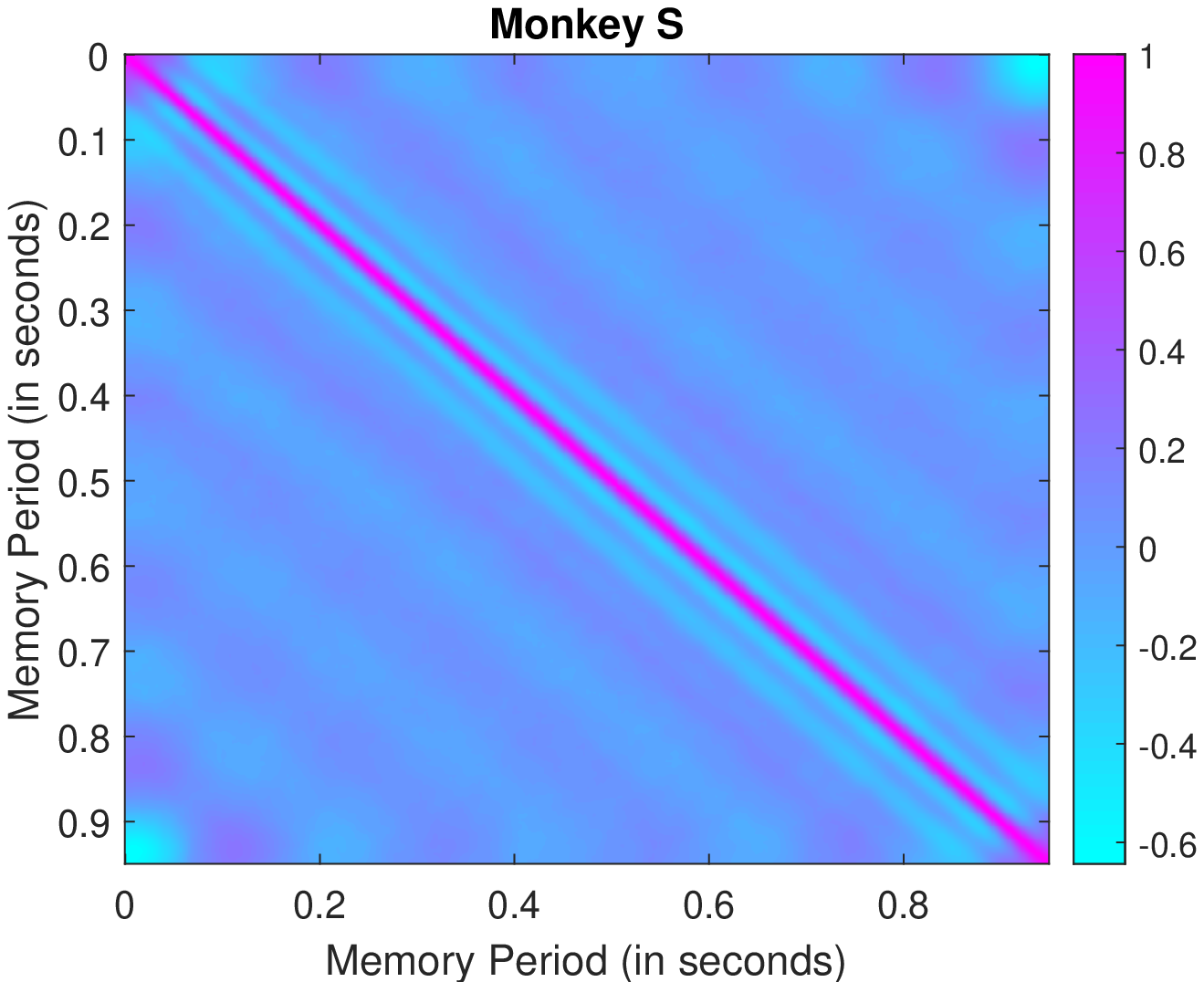}\label{MonkeyS_corrMat}}
\hfil
\subfloat{\includegraphics[scale=0.5]{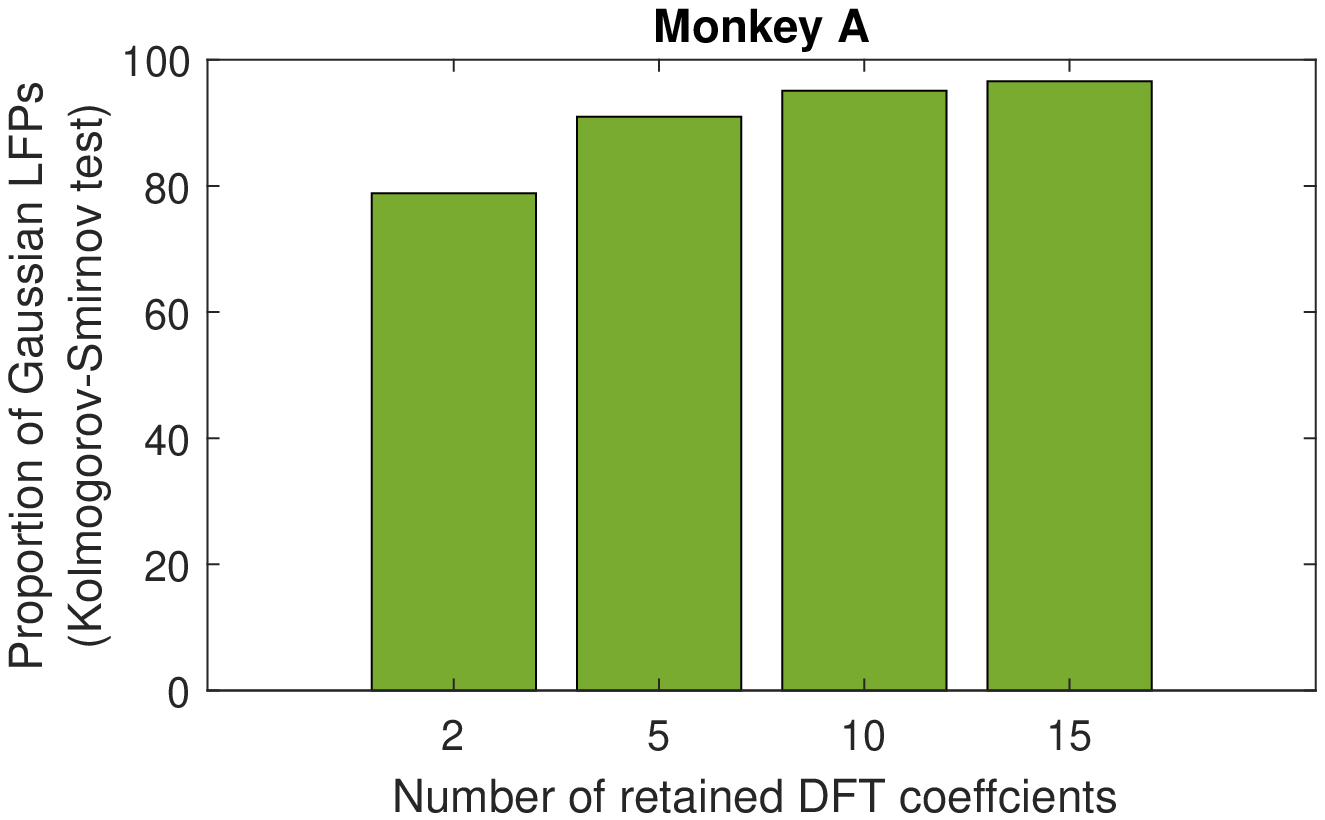}\label{MonkeyA_KS}}
\hfil
\subfloat{\includegraphics[scale=0.5]{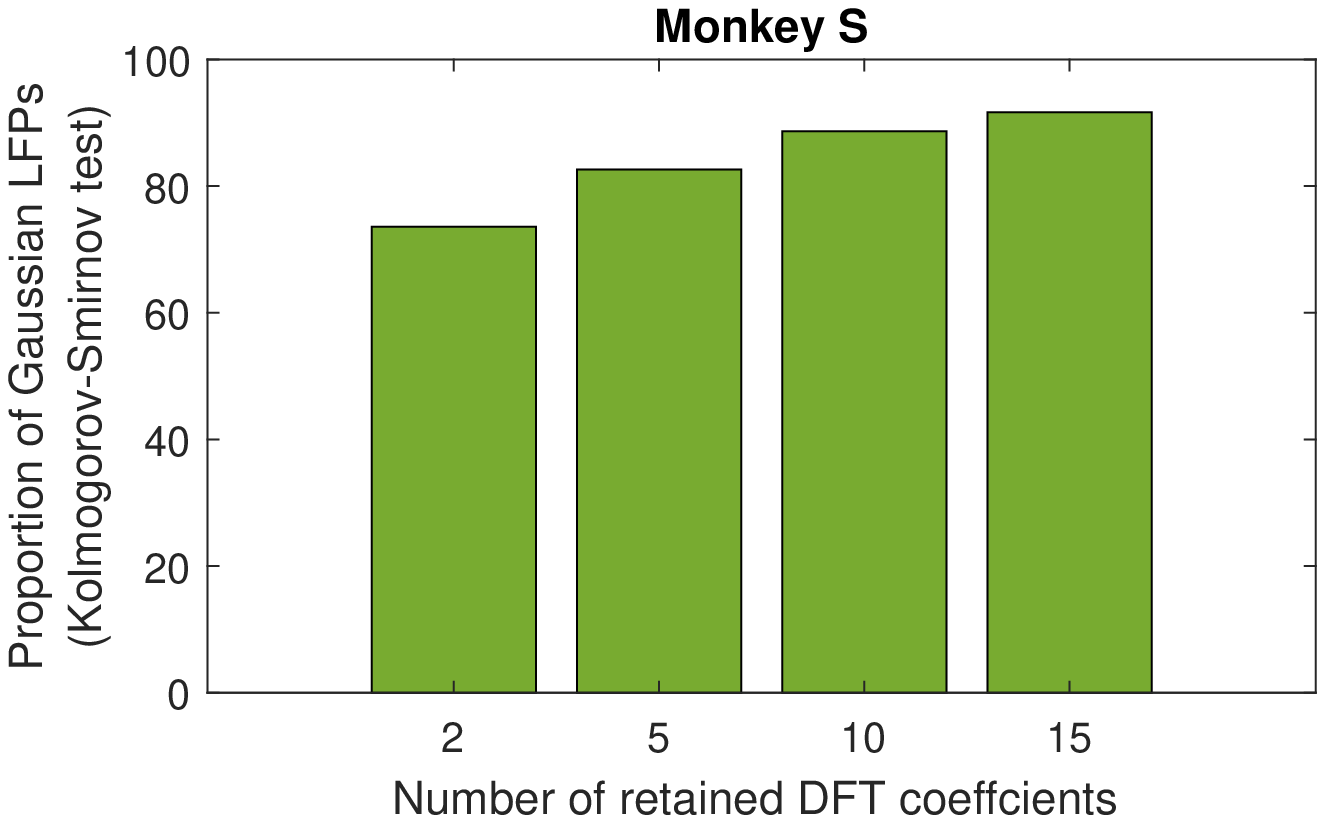}\label{MonkeyS_KS}}
\caption{\RI{Statistical properties of the noise components. The channel with index $10$ (see Fig.~\ref{Monkey_brain} for details on the channel indexing) was arbitrarily selected; same conclusions apply to any other channel. Top row plots: correlation matrices of the noise component, averaged over all trials. The original LFP signals were low-pass filtered with \eqref{eq:truncation} with cut-off frequency $5$ Hz. Bottom row plots: proportion of noise sequences that confirmed the null hypothesis in Kolmogorov-Smirnov goodness-of-fit test with significance level of $0.01$.}}
\label{LFPNoiseAnalysis}
\end{figure}

\RI{The smoothness condition in Pinkser's theorem essentially requires the spectral density of the LFP signals to be a non-increasing function of the frequency; in other words, the structure of the signal should be primarily dominated by low frequency components.
Fig.~\ref{LFPSignalsAndTheirPSDs} depicts multiple single-trial LFP signals (left-hand side plots) as well as their corresponding PSDs (right-hand side plots), before (blue lines) and after low-pass filtering (red lines) using the truncation estimator \eqref{eq:truncation} with cut-off frequency of $5$ Hz (see the caption for more details); the lines are superimposed on the same plots for better visual depiction of the underlying trends in the LFP signals and their PSDs.
We see that in the time domain (left-hand side plots), a strong low frequency component, which becomes even more apparent after low-pass filtering can be identified in both subjects. Correspondingly, the PSD profiles of the LFP signals before low-pass filtering (right-hand side plots) show an exponential decay trend, implying that the smoothness criteria required by Pinsker's theorem are met.}

\RI{To assess the validity of the model \eqref{eq:lfp_model}, we analyse the statistical properties of LFP signal noise.
To extract the noise from the LFP data, we first low-pass filter the noisy LFP signals using the truncation estimator \eqref{eq:truncation}, and then we subtract the reconstructed, smooth signal from the original signal.
We then compute the correlation matrix and quantify the Gaussianity of the LFP noise using the standard Kolmogorov-Smirnov goodness-of-fit (KS-GOF) test \cite{kstest1951}. Recall that the null hypothesis in single-trial KS-GOF test is that the LFP noise values are independent realizations of the standard normal distribution; therefore, before running the test, we normalize the noise sequence by dividing with its standard deviation. 
The results are shown in Fig.~\ref{LFPNoiseAnalysis}.
The top row depicts the correlation matrices of the noise; we can clearly see that the matrices for both subjects are diagonally dominated, indicating that the temporal independence assumption from \eqref{eq:lfp_model} is adequate.
The bottom row plots shows the proportion of all LFP noise sequences that have confirmed the null hypothesis with significance level of $0.01$ for multiple cut-off frequencies in the range $2-15$ Hz.
We observe that the normality assumptions required in the model \eqref{eq:lfp_model} is strongly implied; even in the case of cut-off frequency of $2$ Hz, which corresponds to $L=2$ retained frequencies, the proportion of LFP noise sequences that confirmed the null hypothesis is nearly $80\%$ in both subjects.}

\subsection{Data Centering for Cross-subject Decoding}\label{sec:cross_subject}

\RIII{The feature spaces are non-stationary across time, space and subjects \cite{FLotte_2015, Jayaram_2016}.}
Specifically, the target-conditional distributions, representing the exact same motor intentions are in general different in different subjects even when the experiment is repeated under the exact same conditions.
As a result, a well-performing and reliable decoder trained on one subject will perform poorly when tested on another subject without adequate pre-processing \cite{Jayaram_2016,Lotte_2018}.
Our preliminary investigations have shown that a well-performing decoder, trained using a data set collected from one subject performs no better than a random choice decoder when applied to a test data collected from different subject.

Our remedy to this problem, which is also our main contribution in this paper, is the data centering procedure which relies on the core assumption that the target-conditional distributions corresponding to the same motor intention are functionally related via deterministic component. In our approach, these functional relations are captured via transfer functions; adequate modelling and estimating of the transfer functions is critical for the success of data centering and it is in the focus of the following subsections.

\begin{figure}[h]
\centering
\includegraphics[scale=0.3]{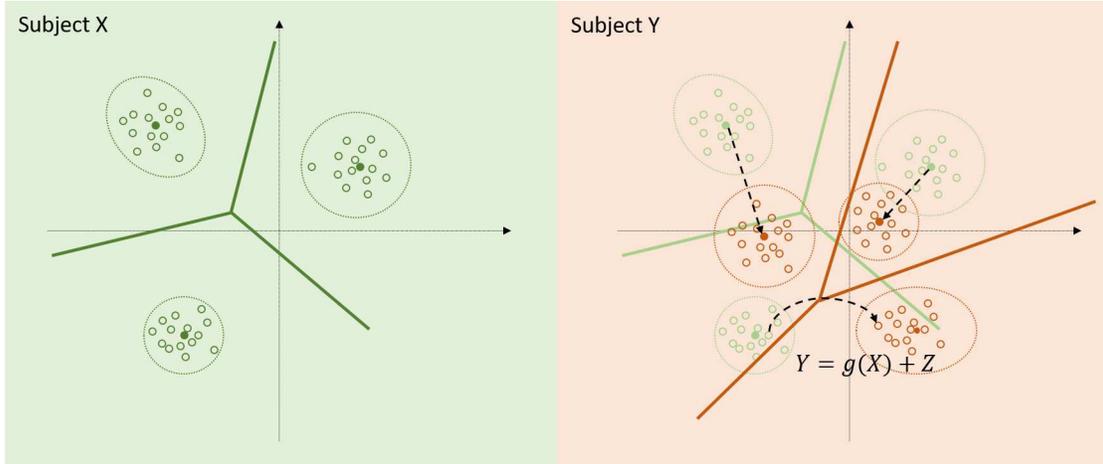}
\caption{Illustration of the data centering procedure in two-dimensional feature space with three target-conditional distributions each of them corresponding to a different motor action. Full circles denote the target-conditional mean vectors, whereas the covariance matrices are represented with dotted circles around the mean vectors. Empty, solid line circles denote individual trials. Solid lines denote decision surfaces. The dashed black lines symbolize the transfer functions, illustrating the transformation of each target-conditional distribution from the feature space of subject X into the feature space of subject Y.}
\label{CenteringEx}
\end{figure}

\subsubsection{Transfer Functions}\label{sec:transfer_functions}
To illustrate the underlying principles of data centering, we fix the following terminology.
Let X denote the subject from which the training data is collected, i.e., the source, and let Y denote the subject where the test data is generated, i.e., the destination. Our goal is to train a decoder using subject X training data, that can reliably decode the subject Y test data.
\RI{Crucial to our method are the following two assumption: (1) the set of performed motor actions is \emph{identical} for both subject X and subject Y, and (2) the subjects perform the motor actions under the \emph{exact same} conditions. Both these assumptions are met in the experiment described in Section~\ref{sec:experiment}: both subjects perform the same memory-guided visual saccades under the same experimental conditions, namely, the target topology on the screen is the same, the subjects are placed on equal distances from the screen, the trials follow the same temporal pattern with equal parameters, the recording chambers were placed in the same part of the PFC for both subjects as shown in Fig.~\ref{Monkey_brain}. 
Provided that these two assumptions are satisfied, we can establish a correspondence between the statistical representation of the motor action set in the feature spaces of the subjects.
Specifically, let $X$ and $Y$ denote the random vectors representing the feature spaces of subject X and Y, respectively; both $X$ and $Y$ live in $\mathbb{R}^n$ where $n\geq 1$ is the dimension of the feature space. $X$ and $Y$ are drawn from a specific class-conditional distribution (we also use the term target-conditional interchangeably) corresponding to a specific motor action (for instance, the visual saccade in direction $\rightarrow$); however, we omit to denote this dependence explicitly to avoid clutter and keep the exposition simple. 
Data centering borrows from transfer learning \cite{TransferL_survey1,Weiss2016ASO} by \emph{a priori} postulating the following (see Fig.~\ref{CenteringEx}): the feature space of subject Y can be viewed as a {functional transformation} of the feature space of subject X. In other words, if $X$ and $Y$ are acquired while subjects X and Y are performing the \emph{same} motor task under the same experimental conditions, we can establish the general relationship: 
\begin{align}\label{eq:XY_general_mapping}
    Y = g(X) + Z.
\end{align}
Here, the invertible and deterministic mapping $g(\cdot)$ relates the feature space representations of the same motor action in two different subjects and is termed \emph{transfer function}; we note that the transfer function $g(\cdot)$ is (in general) \emph{different} for different motor actions, which is illustrated in Fig.~\ref{CenteringEx}.
In addition to the deterministic component, the model \eqref{eq:XY_general_mapping} includes a stochastic, noise-like component, i.e., subject-specific feature noise, represented via the random vector $Z$ which captures the local brain dynamics that further perturbs the feature space of subject Y, independently of $X$.
}

Now, according to \eqref{eq:XY_general_mapping} (see also Fig.~\ref{CenteringEx}), if the transfer functions $g(\cdot)$ for each specific motor intention in the action set are known and if the local feature noise is not too large, the subject X data can be transferred into the feature space of subject Y, as if it was generated there in the first place. This is the core idea of the data centering technique: provided that $g(\cdot)$ is known, the training data set from subject X can be transferred to subject Y by simply applying $g(\cdot)$ to it; the decoder trained using the transformed data set can be subsequently used to decode intended actions by subject Y.

\RI{Data centering is essentially a domain adaptation transfer learning method since it operates over the feature spaces of the subjects with the goal of aligning the corresponding class-conditional distributions for each motor action in a supervised manner. However, unlike conventional domain adaptation methods in non-invasive BCIs where destination data is projected onto common feature space, estimated using several sources, data centering applies transfer functions to map source data onto the destination feature space, essentially combining the advantages of both domain adaptation and ensemble learning.}
\RII{This makes data centering a versatile transfer learning method whose appeal extends beyond cross-subject decoding and can easily be applied to cross-session and inter-cortical decoding to address the temporal and spatial non-stationarity of subject-specific feature spaces without any technical modification of the general model \eqref{eq:XY_general_mapping}.
For instance, let $X$ be a feature vector collected from a subject while performing a given motor action during recording session X; similarly, let $Y$ be a feature vector collected from the same subject performing the same motor action during recording session Y (e.g. the following day). In such case, the general model \eqref{eq:XY_general_mapping} is still valid; the transfer functions $g(\cdot)$ now capture the temporal variability of the subject-specific feature space and data centering can be applied to transfer the training data collected during session X to the feature space characterizing session Y.
In light of this, data centering bears conceptual similarities with the covariate shift method presented in \cite{Sugiyama:2007:CSA:1314498.1390324} to address classification problems with non-stationary temporal data. 
}

\begin{figure*}[h]
\centering
\includegraphics[scale=0.3]{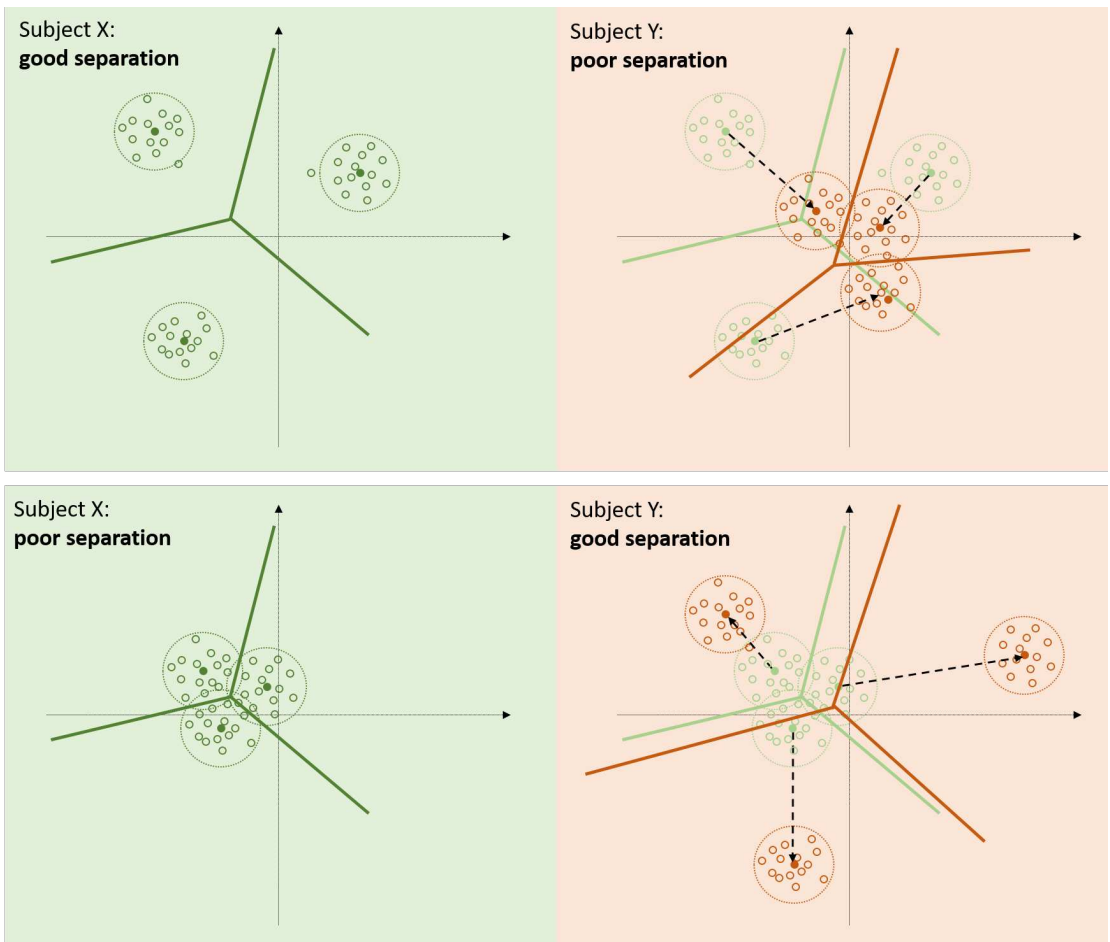}
\caption{\RIII{Impact of the geometry of the feature spaces on data centering. Illustration with two-dimensional features. Top cartoon: the class-conditional distributions of the source subject are well-separated but they are mapped to a destination feature space with poorly separated class-conditional distributions. Bottom cartoon: the class-conditional distributions of the source subject are poorly separated but the data is mapped to a destination feature space with well-separated class-conditional distributions.}}
\label{FeatSpaceMapping}
\end{figure*}

By definition data centering does not require more than one source data set which is one of its main advantages over other domain adaptation methods that operate over common feature space which is estimated using multiple source data sets.
This added flexibility, however, comes at the price of increased susceptibility to the geometry of the destination feature space as illustrated in Fig.~\ref{FeatSpaceMapping}; namely, \RIII{a source data set that performs well when tested on the source subject, might exhibit a considerable drop in performance when transferred to a destination feature space due to poor separation between the class-conditional distributions, see the top cartoon. Oppositely, a source data set that performs poorly on the source subjects might show a significant performance improvement in the destination feature space, see the bottom cartoon.
As it turns out in our experimental investigations, the geometrical aspects of the source and destination feature space indeed play an important role in the success of data centering.
They are however, not the only decisive factor that determines the performance and the transfer function model, the transfer function estimation procedure as well as the subject-specific feature noise will also impact the performance of the source data set in the destination feature space.
}

\subsubsection{Learning Linear Transfer Functions}\label{sec:learnLinTransfer}
As seen from \eqref{eq:XY_general_mapping}, knowledge of the transfer function $g(\cdot)$ is crucial for the success of data centering. Obviously, identifying an adequate model from first principles using neurological reasoning and modelling is a challenging problem as it is unclear in advance how the feature spaces across different subjects in a population can be mapped onto each other. An alternative way to obtain knowledge of the transfer functions is through data-driven modelling and estimation.
In other words, one can resort to data collected from both subjects X and Y and estimate the transfer function relying on either parametric or non-parametric model. In this work, we restrict our attention to simple linear transfer functions: 
\begin{align}\label{eq:XY_linear_mapping}
    Y = HX + Z,
\end{align}
where $H$ is an invertible square matrix. The linear transfer function model might be adequate in situations where the feature vectors follow multivariate normal distributions and where the assumption that the statistical nature of the feature space distributions for specific set of actions should be preserved across subjects is reasonably justified. The advantage of the data-driven estimation approach is in its versatility; as long as there is enough data, one can fit a transfer function for any pair of subjects. In addition, provided that there are enough representatives of the population, the estimated transfer functions form a manifold which can be further used to analyze the general mapping principles for the decision-making part of the PFC across an ensemble of subjects performing the same sets of tasks; such investigations are beyond the scope of this paper.

Estimating models of the form \eqref{eq:XY_linear_mapping} is a well understood problem in statistical signal processing and dynamical systems and one could be tempted to apply popular methods for finding $H$ such as the method of least squares for instance \cite{KayI}. However, in our case there is one critical twist that makes the problem significantly more challenging: since we are mapping feature spaces of two different subjects, no causal input-output relationship can be established between different realizations of $X$ and $Y$ in the corresponding training data sets. In other words, we are unable to tell which specific realization of $Y$ corresponds to a given realization of $X$.
As an alternative, we develop an estimation approach that relies on the first and second-order moments, that is, the mean vectors and covariance matrices of the class-conditional distributions; this leads to a particularly simple estimation approach of the linear model \eqref{eq:XY_linear_mapping}.
Specifically, we assume that the first and second-order moments of $X,Y$ exist and we denote them as $\mathbb{E}(X)=\mu_X$, $\mathbb{E}(Y)=\mu_Y$, $\text{Cov}(X)=\Sigma_X$ and $\text{Cov}(Y)=\Sigma_Y$. Furthermore, we assume that the local, subject-specific noise in subject $Y$ has zero mean, i.e., $\mathbb{E}(Z)=0$ with unknown covariance matrix denoted by $\text{Cov}(Z)=\Sigma_Z$; hence, in the model \eqref{eq:XY_linear_mapping} we have two unknown components: the transformation matrix $H$ and the local noise covariance matrix $\Sigma_Z$.
For reasons that will become apparent shortly, we define the matrix $W_Y$ such that $\text{Cov}(W_YY)=I$; an example would be the standard whitening transformation $W_Y = \Sigma_Y^{-\frac{1}{2}}$.
Using this notation, it is not difficult to show that the following holds:
\begin{align}\label{eq:mean_linear}
    \mu_Y & = H\mu_X,\\\label{eq:cov_linear}
        I & = W_YH\Sigma_XH^TW_Y^T + W_Y\Sigma_ZW_Y^T.
\end{align}
The maximum number of linearly independent equations in the above system is $n+n^2$, \RIII{where $n$ is the dimension of the feature spaces of the subjects}, whereas the total number of unknown parameters is $n^2 + \frac{n(n+1)}{2}\geq n+n^2$ with equality only for the case $n=1$.

In order to solve the above system for the transfer function $H$ in presence of unknown local noise covariance $\Sigma_Z$, we make several simplifying but reasonable assumptions. First, we restricts $H$ to be a diagonalizible matrix. Second, we assume that the Frobenius norm\footnote{The Frobenius norm of a matrix $A$ is $\|A\|=\sqrt{\text{Trace}(AA^T)}$.} $\|W_Y\Sigma_ZW_Y^T\|<1$ which implies that the matrix $W_Y\Sigma_ZW_Y^T$ is diagonally dominated by the identity matrix $I$; in fact, we will assume that the local noise process $Z$ is small in comparison with $Y$ which allows us to use the first order approximation of the Neumann expansion $(I+A)^{\frac{1}{2}} \approx I + \frac{1}{2}A$,
valid for $A$ satisfying $\|A\|\ll 1$.
Third, to obtain well-posed estimation problem for $H$ in presence of unknown $\Sigma_Z$, we need to impose certain structure on $\Sigma_Z$ by incorporating some reasonable intuitions; this is achieved via the parametrization $\Sigma_Z = C(\theta)$ where $\theta$ is unknown parameter vector with $\text{dim}(\theta)\leq n$. Finally we will allow $C(\theta)$ to have all degrees of freedom but we will restrict it to be diagonal, i.e., $\Sigma_Z = \text{diag}(\theta)$; with this final restriction, we assume that the local brain dynamics in subject Y acts on each feature independently.

Taking into account all assumptions stated above it is easy to obtain an estimate of the transfer function $H$. Namely, from \eqref{eq:cov_linear} and using the assumption that $H$ is diagonalizible, we obtain
\begin{align}\label{eq:Hestimation_I}
    H = W_Y^{-1}(I - W_Y\Sigma_ZW_Y^T)^{\frac{1}{2}}\Sigma_X^{-\frac{1}{2}}.
\end{align}
Applying the assumption that the local perturbation process in subject $Y$ is small, we obtain
\begin{align}\label{eq:Hestimation_II}
    H \approx W_Y^{-1}\left(I-\frac{1}{2}W_Y\Sigma_ZW_Y^T\right)\Sigma_X^{-\frac{1}{2}}.
\end{align}
Finally, plugging \eqref{eq:Hestimation_II} in \eqref{eq:mean_linear} and assuming $\Sigma_Z = \text{diag}(\theta)$, we obtain an estimate of $\theta$ as
\begin{align}\label{eq:Hestimation_III}
    \hat{\theta} = 2\text{diag}\left(W_Y^T\Sigma_X^{-\frac{1}{2}}\right)^{-1}(W_Y^{-1}\Sigma_X^{-\frac{1}{2}}\mu_X - \mu_Y),
\end{align}
which is then replaced in \eqref{eq:Hestimation_II} to obtain an estimate for the transfer function $H$.

The reader should note that the above estimation approach is applied for each target-conditional distribution individually; hence, if the number of targets is $8$ as in the eye movement decoding problem, we need to estimate $8$ corresponding linear transformation matrices $H$, one for each target-conditional distribution. In light of this, from \eqref{eq:Hestimation_II} and \eqref{eq:Hestimation_III} we observe that the computation of both $\theta$ and $H$ involves inverse of the target-conditional covariance matrices. In a limited data scenario, we need to make sure that the covariance matrices are well-conditioned such that they can be adequately inverted. This might require reducing the dimension during the feature extraction phase by, for instance, using lower cut-off frequency in \eqref{eq:truncation} or increasing the number of trials per data set using the data bundling technique from \cite{Angjelichinoski2019}, see also Section~\ref{sec:databundling}.
Alternatively, in cases when neither of these two options is applicable, we can resort to the shared covariance matrix instead, which is computed as a linear combination of the target-conditional covariance matrices with weights given by the empirical target priors \cite{Bishop:2006:PRM:1162264}; this approach will be useful later on Section~\ref{sec:datacenteringutility} when we investigate the performance of data centering on imbalanced data sets where one or more targets distributions are poorly represented, i.e., the number of trial is significantly less than the dimension and where further reduction of the feature space dimension is not desirable.

\RI{Compared with the covariate shift method from \cite{Sugiyama:2007:CSA:1314498.1390324} where the authors use weighted importance sampling, which requires knowledge of the underlying class-conditional distributions, our approach has its simplicity as its main advantage; using the linear transfer function model \eqref{eq:XY_linear_mapping} and relying on the simplifying assumptions regarding the structure of the transfer function $H$ and the local perturbation process $Z$, we were able to derive an approximate, closed form solution for $H$ using only the mean vectors and covariance matrices of the source and destination features $X$ and $Y$ which can be easily estimated using training data.
This approach, on the other hand, is somewhat reminiscent to the ``frustratingly easy'' domain adaptation method introduced in \cite{Sun:2016:RFE:3016100.3016186}. However, unlike the method from \cite{Sun:2016:RFE:3016100.3016186}, which is unsupervised and uses all available destination data (including the test data) to estimate the second-order moments, data centering is a supervised method and it is not allowed to see the destination data used for testing; more details can be found Section~\ref{sec:datadrivendatacent} and in Fig.~\ref{DataCentering}.
}
It should be noted however that the transfer function $g(\cdot)$ need not be linear in general.
In fact, it is more reasonable to assume more general, non-linear mapping between the feature spaces and use sophisticated methodology for estimating the non-linear transfer functions using deep neural networks; such modelling and estimation approaches are part of our ongoing research and are beyond the scope of this paper.

\subsection{Data Preparation and Processing}\label{sec:DataPrepar}

\subsubsection{Data Description}\label{sec:datadescription}

The data analyzed in the rest of the paper was first reported in \cite{Markowitz18412} and subsequently used in \cite{Markowitz11084,Angjelichinoski2019}. 
The subjects were trained across several days and recording sessions.
As also described in Section~\ref{sec:experiment}, over the course of the experiment, the positions of the electrodes were gradually advanced into the PFC; each unique configuration of electrode depths, described by $32$-dimensional vector is referred as EDC, see Section~\ref{sec:experiment}. The experiment was performed over a total of $34$ and $55$ EDCs for Monkey A and Monkey S, respectively, during which multiple trials were collected; the indexing of the EDCs used here reflects the temporal progression of the experiment; namely, the trials for EDC-$i$ were collected earlier in time than the trials for EDC-$j$ if $j>i$. 
The total number of trials across all EDCs is $3922$ for Monkey A and $13064$ for Monkey S.
As detailed in \cite{Markowitz18412,Angjelichinoski2019}, the experiment for Monkey S started before the electrodes penetrated the surface of the PFC and the first $14$ EDCs and recordings were take while some (or all) of the electrodes were still outside the PFC.
The respective averages are $\approx 90$ and $\approx 250$ trials per EDC for Monkey A and Monkey S, respectively (with the exception of EDC-$6$ in Monkey A for which a total of $827$ trials were collected across $10$ recording sessions).
Given the dimension of the feature space which easily surpasses $100$ (see \cite{Markowitz18412,Angjelichinoski2019}), we conclude that the number of trials for each individual EDC is insufficient to train a reliable decoder.


\subsubsection{Data Bundling}\label{sec:databundling}
To obtain data sets of sufficient size, we apply the trial bundling technique from \cite{Angjelichinoski2019} based on the Euclidean proximity of the EDC vectors using the following reasoning: similar EDCs, i.e., EDCs whose depth vectors are close in Euclidean sense, generate similar feature spaces with similar target-conditional distributions.
The algorithm operates as follows. First, we define \emph{clustering window} as the minimum number of trials per EDC data set, i.e. data cluster. Second, we fix the \emph{concurrent EDC} at which we want to create data set of sufficient size. Then, we begin appending trials for the concurrent EDC. We first append the trials from the concurrent EDC; if the number of trials is less than than the clustering window, we start appending trials from the EDCs which are closest to the concurrent EDC in Euclidean sense. The algorithm terminates when the total number of trials fills up the clustering window. 

\subsubsection{Feature Space Formation}
Once the data sets of sufficient size have been formed, they are processed using the feature extraction procedure described in Section~\ref{sec:pinsker}. Feature extraction is performed on a per channel basis; namely, for each trial, the LFP signals from each individual channel are transformed into frequency domain and they are low-pass filtered using the truncation estimator \eqref{eq:truncation}. The total number of retained (complex) Fourier coefficients per channel is $L$ (corresponding to the DC component and $L-1$ lowest frequencies). We use rectangular representation to store the complex DFT coefficients via their sine and cosine components; hence, the dimension of the number of features per channel is $2L-1$ (the DC component is real, so only one coefficient is stored). These spectral representations are then concatenated across the channels to form one large feature vector of dimension $n=32\cdot(2L-1)$. 


\begin{figure}[h]
\centering
\includegraphics[scale=0.375]{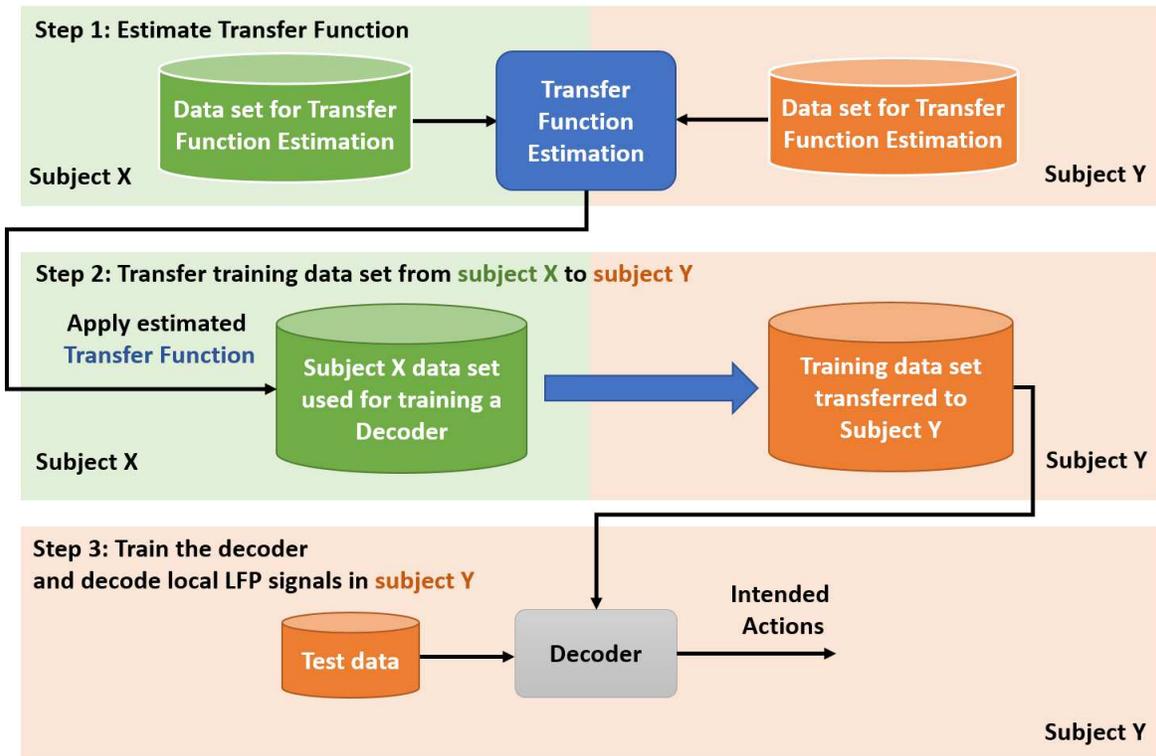}
\caption{Block diagram illustrating data-driven data centering where transfer functions are estimated using the available data from subject X and Y. Following the notation convention, the objective is to decode intended motor actions in subject Y using a decoder trained with subject X data. The procedure follows three steps. Step 1: the transfer function is estimated using dedicated data sets from both subject X and Y. Step 2 (the data centering step): the estimated transfer function is applied to the training data from subject X; after this, the subject X data has been transferred, i.e., centered to subject Y. Step 3: the centered data is used to train a decoder which is then used to decode intended motor actions from unseen test data in subject Y.}
\label{DataCentering}
\end{figure}

\subsubsection{Data-driven Data Centering}\label{sec:datadrivendatacent}
\RI{For cross-subject studies, the next step is to define the source and destination data sets for transfer function estimation, the source data set used for cross-subject training and the destination data set used for testing, see Fig.~\ref{DataCentering}. For the destination (subject Y) this is simple to do: namely, we randomly pull a holdout test set from each EDC data set. The remaining trials in the data set are used for estimating the transfer function. For instance, if we use leave-one-out cross-validation, we randomly pull out single trial from the subject Y data set leaving the remaining trials for transfer function estimation. For the source (subject X) we form two data subsets. Namely, we define a real positive number $\alpha\leq 1$, representing the proportion of trials which are randomly pulled from the source data set. The source data set is then sampled independently twice to form the subset used for transfer function estimation and the subset which is used to train the cross-subject decoder after data centering. This might lead to partial overlap between these two subsets as some trials will likely be pulled out for both transfer function estimation and cross-subject decoder training; this will certainly occur when $\alpha>0.5$. An alternative to this approach is to split the subject X data set into two disjoint subsets following predefined ratio; one subset (e.g. $30\%$ of the trials) is for transfer function estimation while the other one (e.g. the remaining $70\%$ of the trials) is used for cross-subject decoder training. We have tested both approaches, with the first one providing superior performance over the second for limited data sets; hence, here we only report the cross-subject decoding results using the first approach where subject X data set is sampled independently for transfer learning and cross-subject training.}

\subsubsection{Decoding}
The final block in the chain is the actual decoder. Due to the limited amount of training data, prior studies have shown that linear discriminant analysis (LDA) is an adequate decoder. In fact, recent evaluations, partially reported in \cite{Angjelichinoski2019}, have confirmed that in subject-specific studies LDA outperforms other standard classifiers, such as quadratic discriminant analysis (QDA) decoder, logistic regression and support vector machines (SVM). The limited amount of data has so far been the main obstacle for applying sophisticated non-linear methods, such as deep neural networks for classification. Due to its proven robustness for the problem at hand, we also use LDA decoder in the cross-subject studies.
We note that LDA essentially uses the distance of the test feature vector to the means of the target-conditional distributions as a discriminant metric. Hence, the subject-specific decoding performance is an indicator of the geometry of the feature spaces of the subject: higher decoding performance indicates better separation between the target-conditional distributions and vice versa.

\section{Results}\label{sec:evaluations}

In Section~\ref{sec:benchmark}, we first look at the subject-specific performance when the decoder for each subject is trained using its own, subject-specific data and use these results as a benchmark in our cross-subject studies. 
Section~\ref{sec:datacenteringutility} presents the results obtained from the cross-subject studies drawing preliminary conclusions regarding the achievable cross-subject decoding performance. Last but not least, Section~\ref{sec:CenteringImbalanced} demonstrates an obvious benefit from data centering in the case of imbalanced training data sets over \RIII{standard, sampling-based methods for learning from imbalanced data sets \cite{He2009}}; this particular study implies the practical potential of data centering in neural prostheses for restoring lost motor functions in subjects with chronic disabilities.

\subsection{Subject-specific Analysis: Benchmark Decoding Performance}\label{sec:benchmark}

\begin{figure}[h]
\centering
\subfloat[Monkey A]{\includegraphics[scale=0.55]{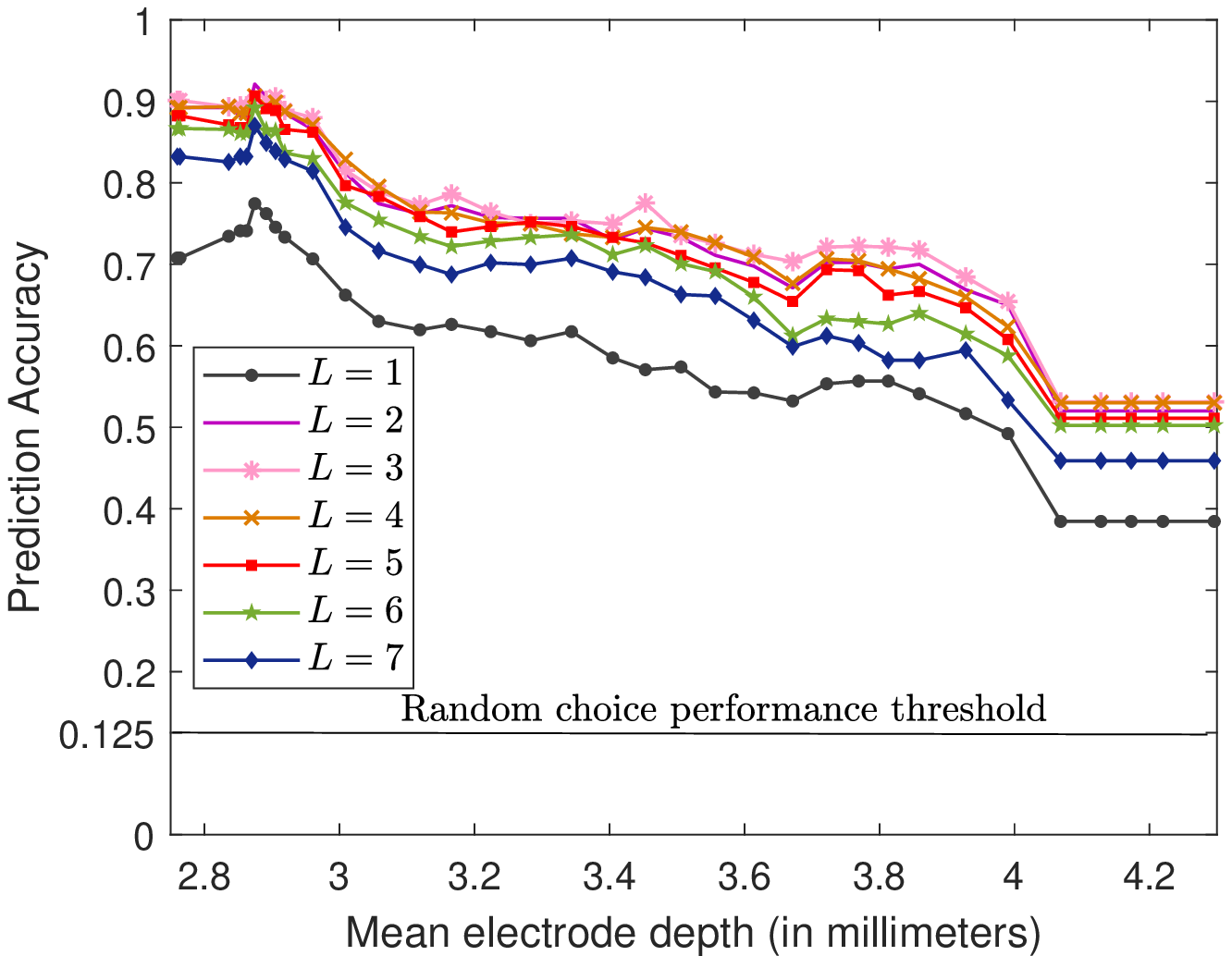}\label{results1a}}
\hfil
\subfloat[Monkey S]{\includegraphics[scale=0.55]{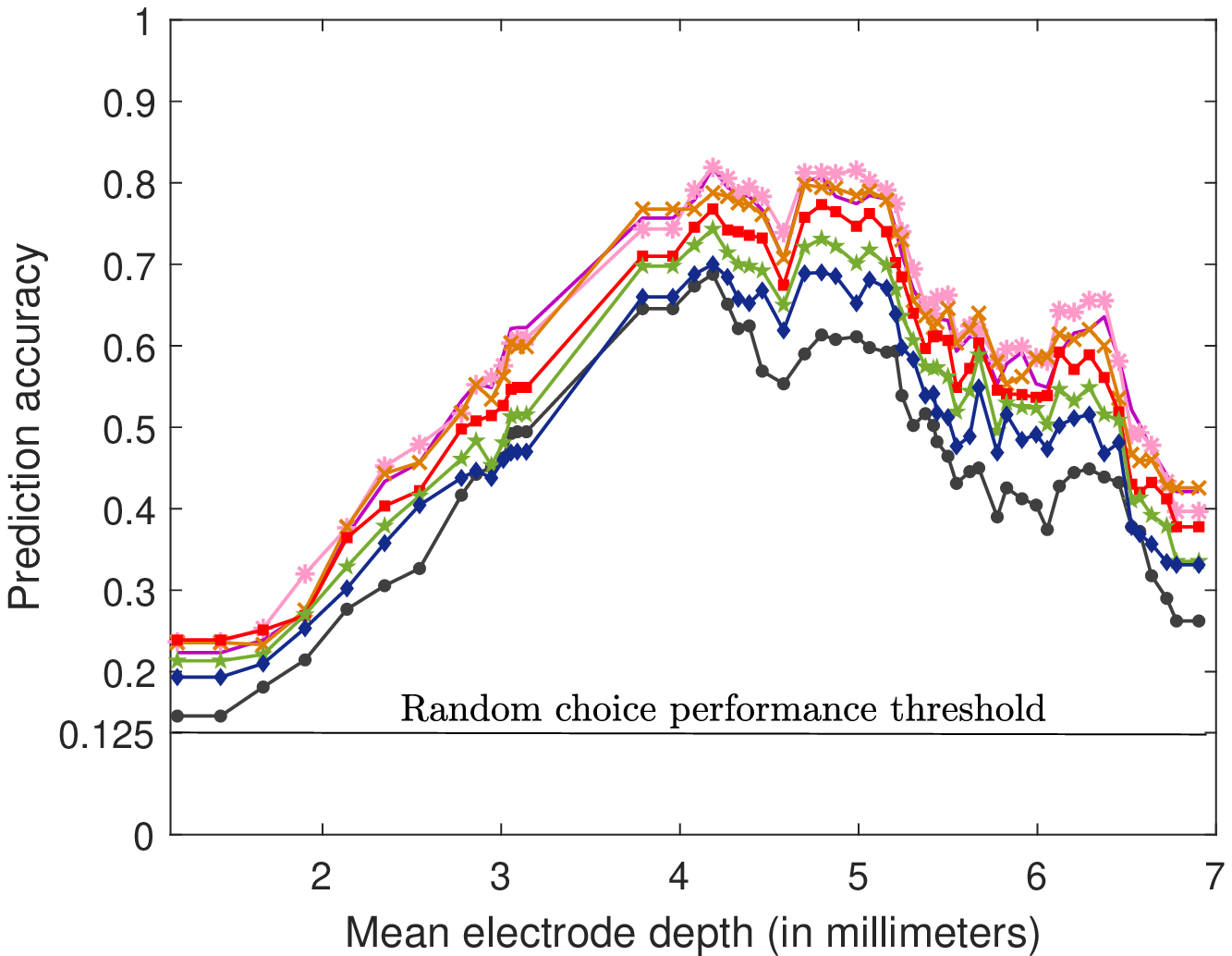}\label{results1b}}
\caption{Subject-specific decoding performance across cortical depths. The horizontal axis gives the mean depth of each array configuration computed as a simple mean of the corresponding EDC vector with respect to the initial positions of the electrodes (see Section~\ref{sec:experiment}). The legend in (a) applies to (b) as well. We use the first $T = 650$ time samples from the memory period \cite{Angjelichinoski2019}. Multiple cut-off frequencies, ranging from $0$ ($L=1$) to $10$ Hz ($L=7$) are considered. The data clustering window is $900$ trials per EDC data cluster, or between $112$ and $113$ per target (see Section~\ref{sec:databundling}).}
\label{results1}
\end{figure}

Fig.~\ref{results1} shows the performance of the LDA decoder trained using Pinkser's features, obtained using \eqref{eq:truncation}. The performance has been evaluated as a statistical average using leave-one-out cross-validation during which we also optimized the value of $T$, i.e., the number of time-domain LFP samples acquired from the memory period.\footnote{As previously reported in \cite{Angjelichinoski2019}, the first half of the memory period, immediately after the target light is switched off, is the most informative; hence, the optimal value of $T$ is between $550$ and $650$ LFP samples per channel.}
We evaluated the performance of the decoder for several values of $L$, i.e., number of retained frequency components; namely, we varied $L$ from $1$ (when only the DC component, i.e., the mean value of the LFP signal is kept) to $7$. Interestingly, the decoder achieves fairly good decoding accuracy compared to random choice decoder ($12.5\%$) even when $L=1$. The decoder achieves its best performance for $L$ between $2$ and $4$; however, for larger $L$, i.e., larger feature spaces the performance of the decoder deteriorates as can be seen from the curves corresponding to $L=5,6,7$.

\RI{The results suggest that the information relevant for the decoding of the memory-driven visual saccade intentions is stored in the DC component and the first frequency component, which for sampling frequency of $\nu_S=1$ kHz corresponds to cut-off frequency of $2.5$ Hz; the next few frequencies, within the first $5$ Hz band provide additional degrees of freedom and can further enhance the decoding accuracy, albeit marginally as evident in Fig.~\ref{results1}. We conclude that the relevant information that determines the dynamics of the memory-driven decision making process for motor intentions in LFPs is stored below $5$ Hz, in the delta band.
This is an interesting finding since the low frequency oscillations in the delta band are conventionally associated with slow-wave sleep stages when no conscious functions occur. In other words, as pointed out in \cite{Nacher15085}, the role of delta band oscillations in decision making dynamics during motor responses has been largely unknown due to lack of investigations. However, it has been observed in \cite{Nacher15085} that the synchronous activity in the delta band is an important contributing factor when large-scale, distant cortical networks coordinate motor responses. Our prior findings in \cite{Angjelichinoski2019} as well as the results presented in Fig.~\ref{results1} further emphasize the importance of delta band oscillations in memory-driven motor response dynamics.}

\subsection{Cross-Subject Analysis}\label{sec:datacenteringutility}

As outlined in the above discussion, the first two DFT coefficients carry virtually all information pertinent to the decoding memory-driven motor intentions from LFP data. Therefore, for our cross-subject studies, we fix $L=2$.

\begin{figure}[h]
\centering
\subfloat[Source is Monkey S, destination is Monkey A]{\includegraphics[scale=0.55]{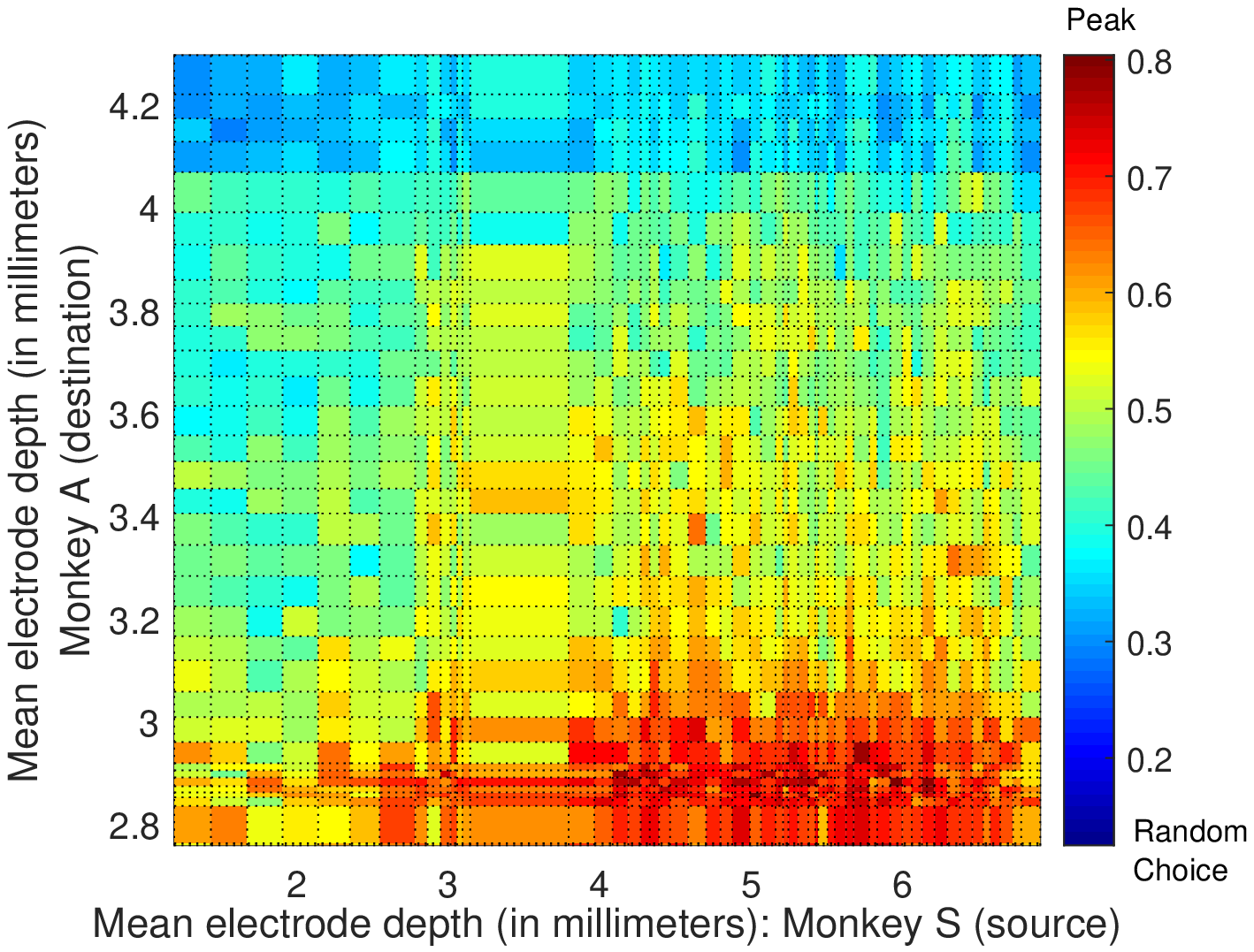}\label{results3a}}
\hfil
\subfloat[Source is Monkey A, destination is Monkey S]{\includegraphics[scale=0.55]{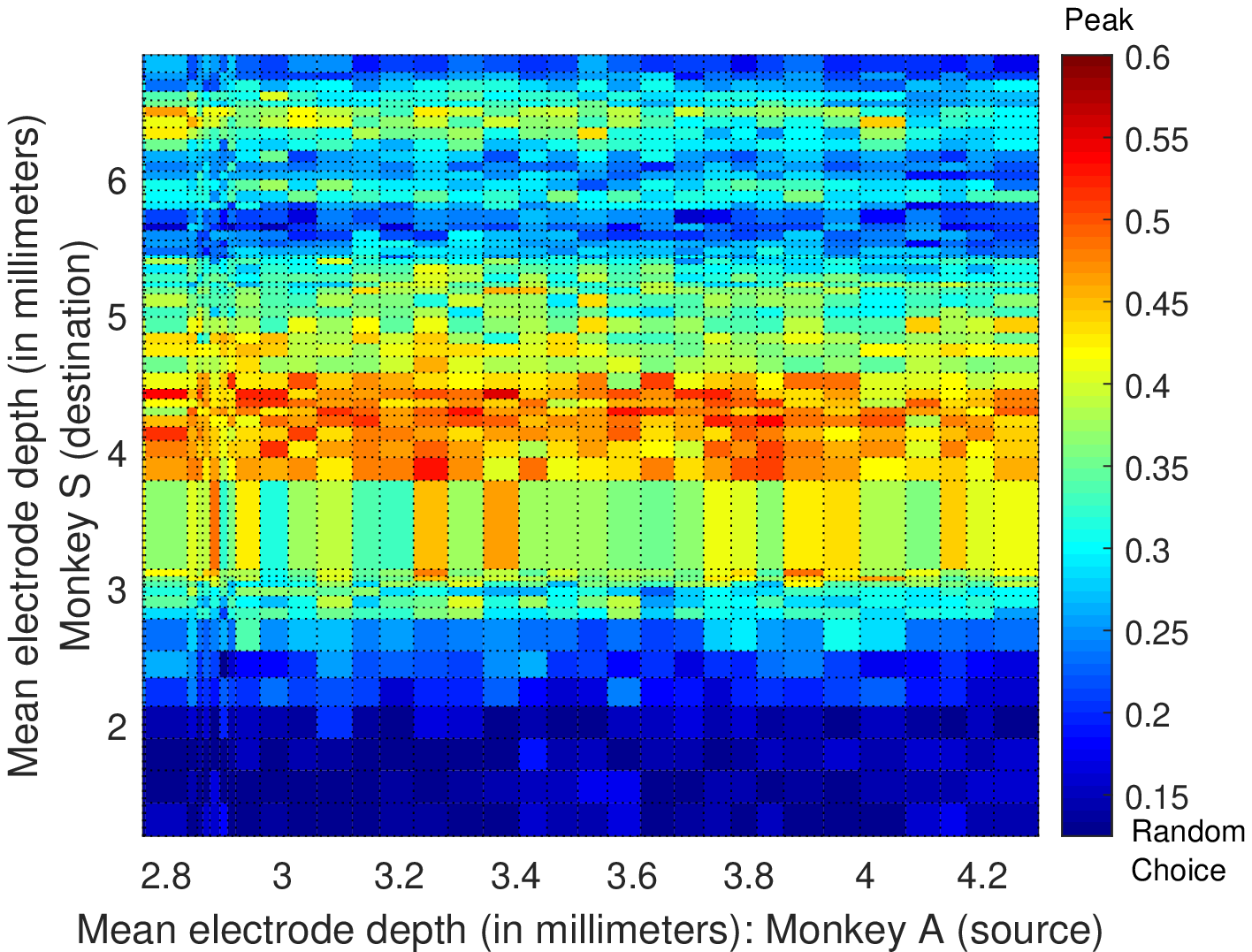}\label{results3b}}
\caption{Cross-subject decoding performance across cortical depths. The axes give the mean depth of each array configuration computed as a simple mean of the corresponding EDC vector with respect to the initial positions of the electrodes (see Section~\ref{sec:experiment}). \RIII{The decoding performance is color-coded; see the colorbars on the right of each plot for the specific color-coding of the prediction accuracy.}  We use the first $T = 650$ time samples from the memory period \cite{Angjelichinoski2019} and features are extracted using \eqref{eq:truncation} with $L=2$. The data clustering window is $900$ trials per EDC data cluster (see Section~\ref{sec:databundling}). We sample the source data set independently with $\alpha=1$ to form the subsets for transfer function estimation and training (see Section~\ref{sec:datadrivendatacent}).}
\label{results3}
\end{figure}
\begin{figure}[h]
\centering
\subfloat[Source is Monkey S (EDC-$32$), destination is Monkey A]{\includegraphics[scale=0.55]{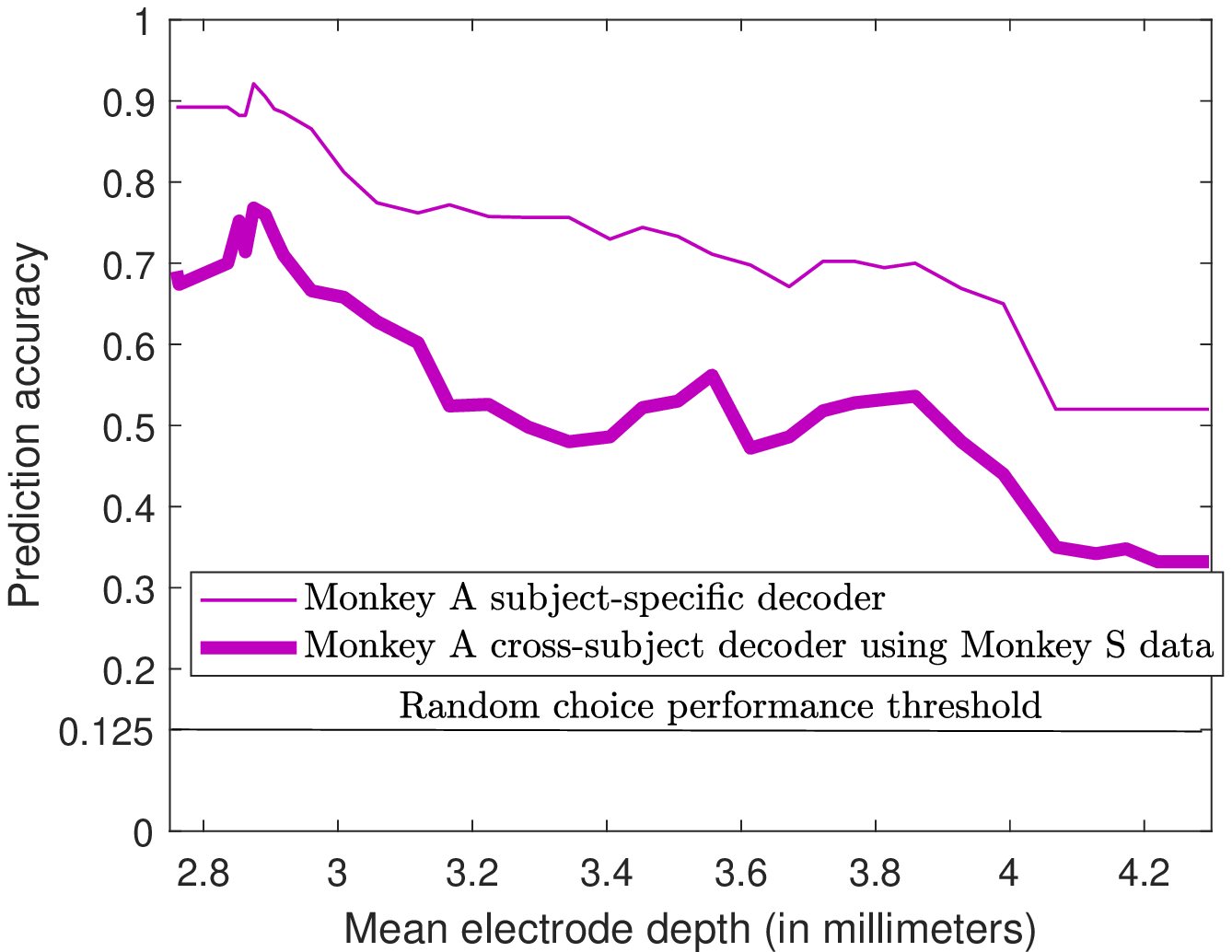}\label{results2a}}
\hfil
\subfloat[Source is Monkey A (EDC-$15$), destination is Monkey S]{\includegraphics[scale=0.55]{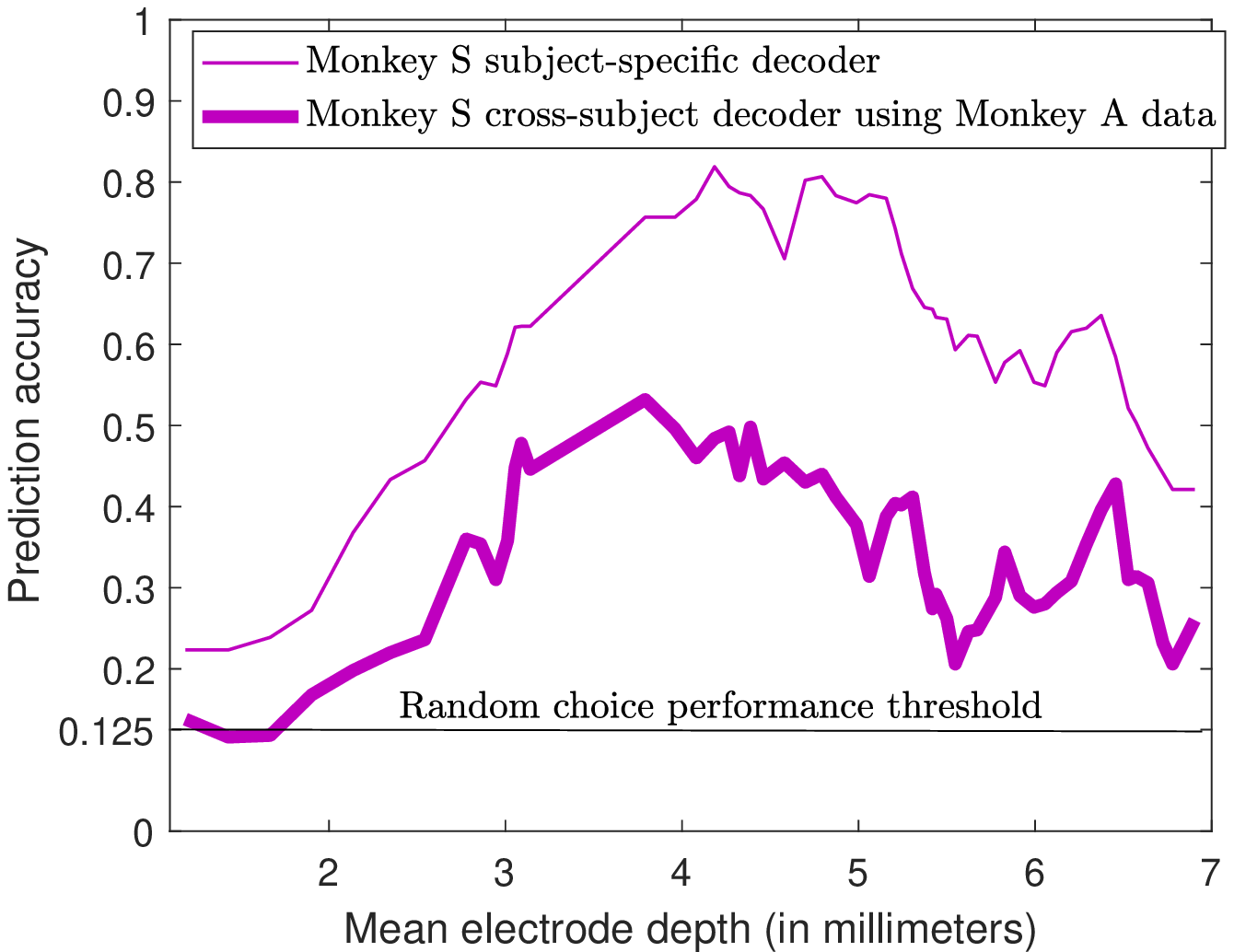}\label{results2b}}
\caption{Cross-subject decoding performance with respect to an arbitrary source EDC data set. The horizontal axis gives the mean depth of each array configuration computed as a simple mean of the corresponding EDC vector with respect to the initial positions of the electrodes (see Section~\ref{sec:experiment}). The source data sets are denoted in the captions of the respective plots. We use the first $T = 650$ time samples from the memory period \cite{Angjelichinoski2019} and features are extracted using \eqref{eq:truncation} with $L=2$. The data clustering window is $900$ trials per EDC data cluster (see Section~\ref{sec:databundling}). We sample the source data set independently with $\alpha=1$ to form the subsets for transfer function estimation and training (see Section~\ref{sec:datadrivendatacent}).}
\label{results2}
\end{figure}
We begin our analysis by first applying data centering between Monkey A and Monkey S across all recording EDCs and compare the resulting performance with the benchmark performance reported in Section~\ref{sec:benchmark}.
The results are shown in \RIII{Figs.~\ref{results3} and~\ref{results2}}.
The two-dimensional plots in Fig.~\ref{results3} depict the cross-subject decoding performance for all available EDC data sets using color-coding; namely, each available EDC data set for subject X is applied to all available data sets for subject Y. The purpose of this analysis is to investigate the performance of data centering for cross-subject decoding across different cortical depths.
Fig.~\ref{results2} compares the cross-subject decoding performance of a single, arbitrarily chosen source EDC from Fig.~\ref{results3} along with the benchmark decoding performance for the destination.

We draw several important conclusions from these results. 
\RII{First, from both Fig.~\ref{results3} and~\ref{results2} we see that the cross-subject decoding is consistently dominated by the locally achievable subject-specific decoding performance when the available data is used for local training instead of transfer learning.
The results suggest that if enough training data is available locally and all target-conditional distributions are equally represented in the feature space, it is recommendable to apply subject-specific training instead of cross-subject decoding via data centering.
This result is intuitively expected and can be explained through the linear transfer function model \eqref{eq:XY_linear_mapping}. Specifically, it can be shown that in the special case when the destination feature noise component $Z$ is zero, data centering essentially maps the mean vectors of the source target-conditional distributions onto the mean vectors of the respective destination target-conditional distributions, see also Figs.~\ref{CenteringEx} and \ref{FeatSpaceMapping}. Since the LDA classifier discriminates based on the Euclidean proximity of the test feature vector to the target-conditional mean vectors, we conclude that best we can anticipate from the linear transfer function model \eqref{eq:XY_linear_mapping} followed by LDA is to replicate the subject-specific performance. The presence of the local feature perturbations as well as the approximations we made in Section~\ref{sec:datadrivendatacent} in order to derive the closed form estimator \eqref{eq:Hestimation_III} additionally limit the cross-subject decoding, creating the observed gap in performance. We conclude that more sophisticated transfer function model might be necessary to further boost the cross-subject decoding performance, which is outside the scope of the current paper.}
Nevertheless, the actual cross-subject decoding performance is also consistently higher and often substantially higher than a random choice decoder, peaking at around $80\%$ in Monkey A, see Fig.~\ref{results3a}; to the best of our knowledge, this work is the first to report such reliable cross-subject decoding accuracy using electrophysiology and we attribute the performance to the data centering method.
\RIII{Next, we observe that several Monkey S EDC data sets achieve higher cross-subject decoding performance when trained in Monkey A feature space after centering, than when trained on Monkey S itself. 
As discussed in Section~\ref{sec:transfer_functions} and illustrated in Fig.~\ref{FeatSpaceMapping}, this is an anticipated result since the target-conditional mean vectors in Monkey A feature space are better separated especially at superficial depths near the surface of the PFC which is indicated by the subject-specific decoding performance curves in Fig.~\ref{results1}. Hence, we are mapping poorly separated Monkey S data onto better separated Monkey A feature space. Conversely, well-performing Monkey A data transferred to Monkey S suffers from performance degradation since the class-conditional mean vectors in Monkey S data are poorly separated as compared to Monkey A feature space.
The geometrical aspects of data cantering also appear to be the dominant reason why the best cross-subject decoding performance only loosely depends on the source data set and most of the cross-subject performance gain is concentrated in a narrow strip of destination EDCs near the surface of the PFC in both Monkeys.\footnote{\RIII{Recall that the first $14$ EDCs in Monkey S were located outside the PFC and the actual penetration into the surface occurs after the mean electrode depth increases beyond $3$ millimeters with respect to the initial position; see \cite{Markowitz18412,Angjelichinoski2019} for more details.}}}

\subsection{Data Centering for Learning from Imbalanced Data Sets}\label{sec:CenteringImbalanced}

\begin{figure}[h]
\centering
\subfloat[Both classes are well-represented]{\includegraphics[scale=0.225]{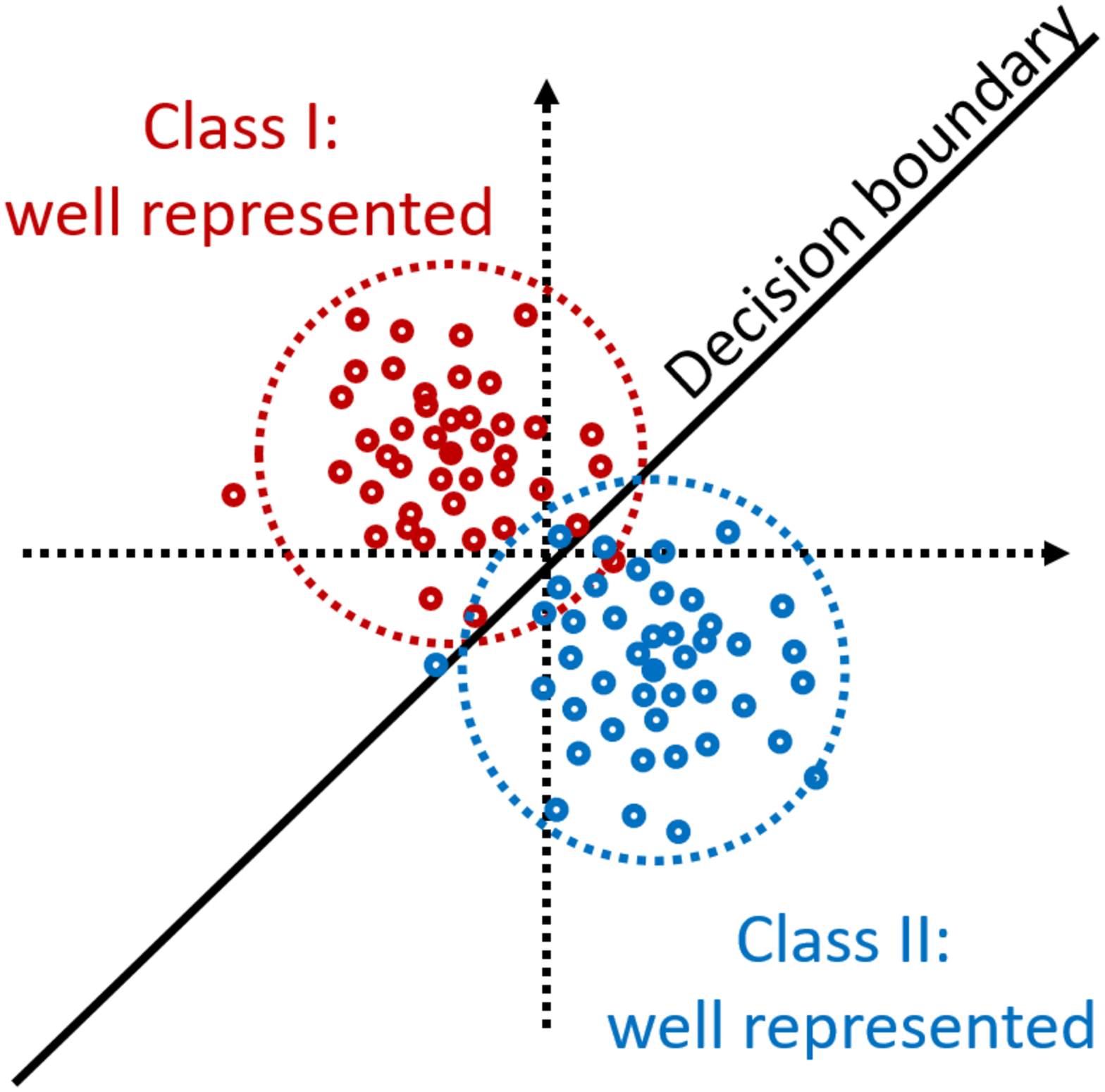}\label{cartoon3a}}
\hfil
\subfloat[One class is under-represented]{\includegraphics[scale=0.225]{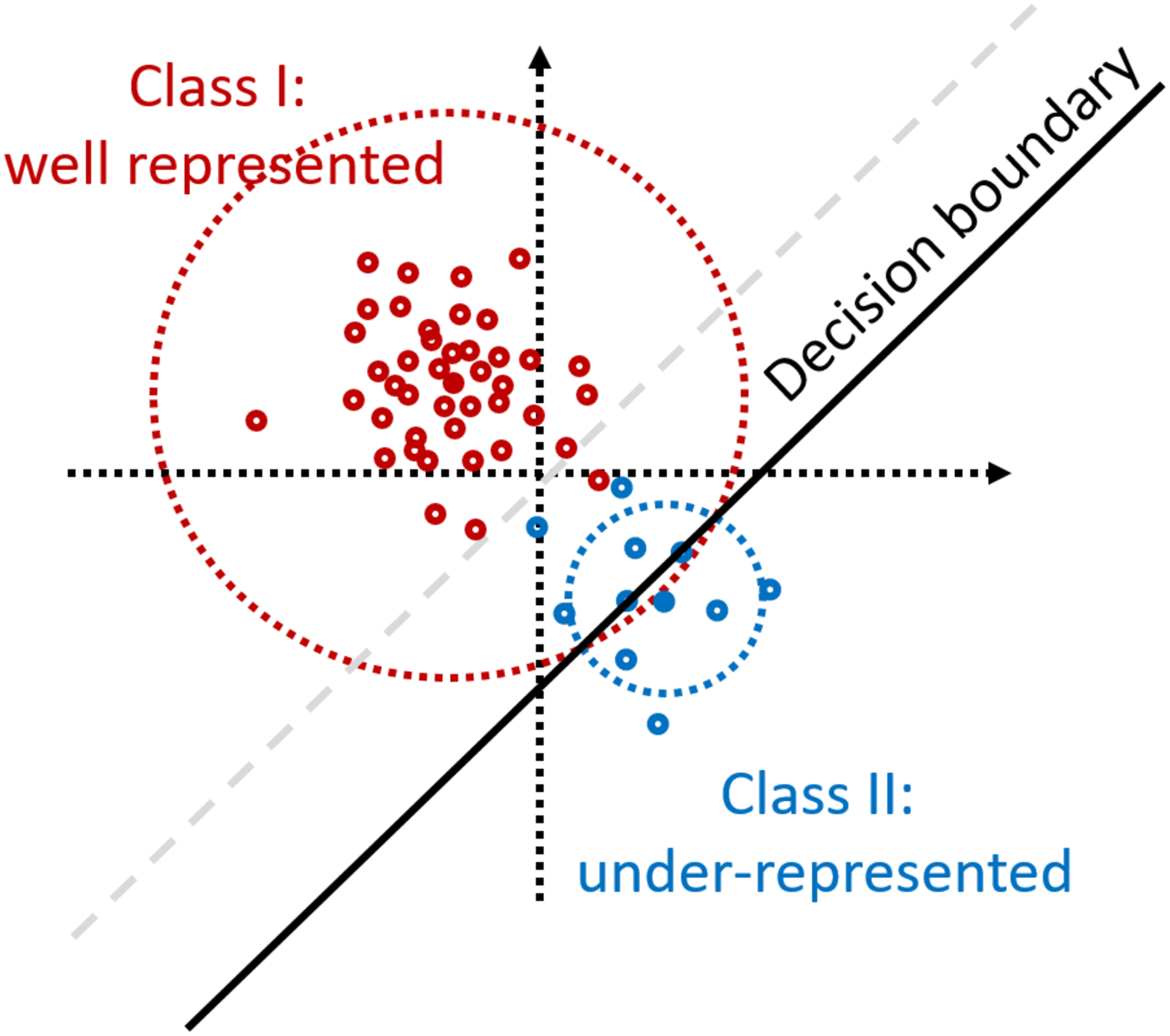}\label{cartoon3b}}
\caption{The impact of imbalanced data in classification problems. Balanced data sets provide fair representation of classes in the feature space and a decoder can be trained reliably as in (a). If one or more classes are under-represented as in (b), the decoder's decisions will shift in favor of the well-represented classes.}
\label{cartoon3}
\end{figure}

\begin{figure}[h]
\centering
\includegraphics[scale=0.225]{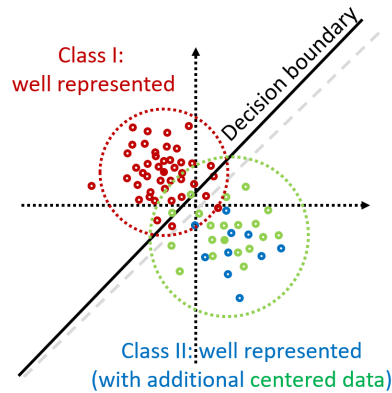}
\caption{Two-dimensional illustration of data centering applied to imbalanced data sets. The imbalance is fixed by bringing data for the under-represented class from different subject (source) after applying data centering.}
\label{cartoon3c}
\end{figure}

One of the most important objectives of cross-subject BCIs is the restoration of lost motor functions in subjects with chronic disabilities; this neurological problem can be partially studied within the framework of learning from imbalanced data sets \cite{He2009}. As an illustration, we consider the following hypothetical scenario. Specific motor functions and actions, such as eye movement directions, in the feature space of a healthy subject (i.e., subject X) will be equally represented; the subject can perform all of the functions into consideration and therefore, a BCI algorithm can easily learn their corresponding feature space representations and decode reliably intended actions as illustrated in two dimensions in Fig.~\ref{cartoon3a}.
On the other hand, a subject that has partially or completely lost the ability to perform one or more of the considered functions (i.e., subject Y) will produce imbalanced feature space where the lost functions are very poorly represented. Hence, a BCI trained over the imbalanced data set will result in poor detection of intended actions from the poorly represented functions, as illustrated in Fig.~\ref{cartoon3b}. A straightforward alternative would be to use a standard sampling technique for dealing with the imbalance between classes such as random oversampling or random under-sampling \cite{He2009}.

Our main question here is whether data centering can leverage the bio-physiological relations and similarities of the LFP data between the two subjects and help to solve the imbalanced data problem, beyond what's achievable with standard techniques. Specifically, we investigate whether we can exploit training data from the well-represented feature space of subject X for the under-represented targets in the feature space of subject Y, as depicted in Fig.~\ref{cartoon3c}.
For illustration purposes, we consider simplified binary hypothesis testing problem in which we pick only two targets, i.e., eye movement directions; then, one of the targets is randomly under-sampled to create artificial imbalance between the targets.
We only consider Monkey S on Monkey A cross-subject decoding with EDC-$6$ data set as destination; the same qualitative conclusions apply in the opposite direction.

\begin{figure}[h]
\centering
\includegraphics[scale=0.7]{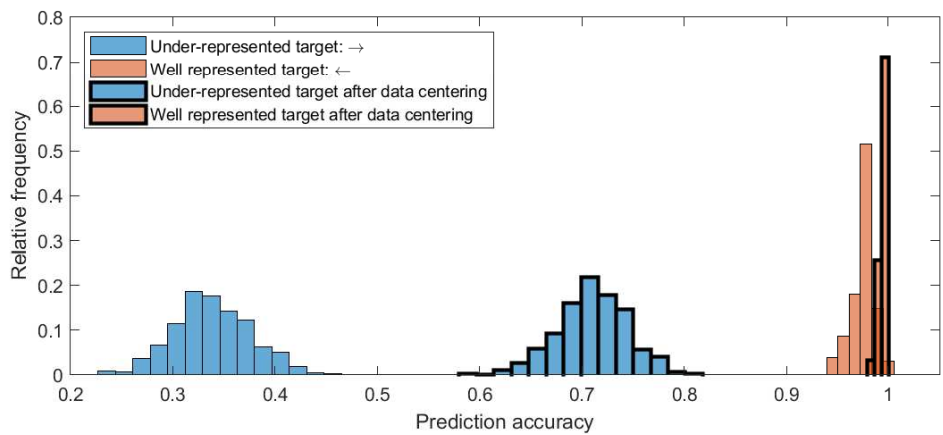}
\caption{Histograms of the cross-subject decoding performance per target with imbalanced data set and between-class imbalance of $100$. Two directions are considered ($\rightarrow$ and $\leftarrow$). EDC-$10$ in Monkey S is source, EDC-$6$ in Monkey A is destination. The histograms are generated using $1000$ randomly-formed subsets for the under-represented target. $T = 650$, $L=2$ and $\alpha=1$.}
\label{histograms}
\end{figure}

\begin{figure}[t]
\centering
\subfloat[EDC-$10$ as source]{\includegraphics[scale=0.55]{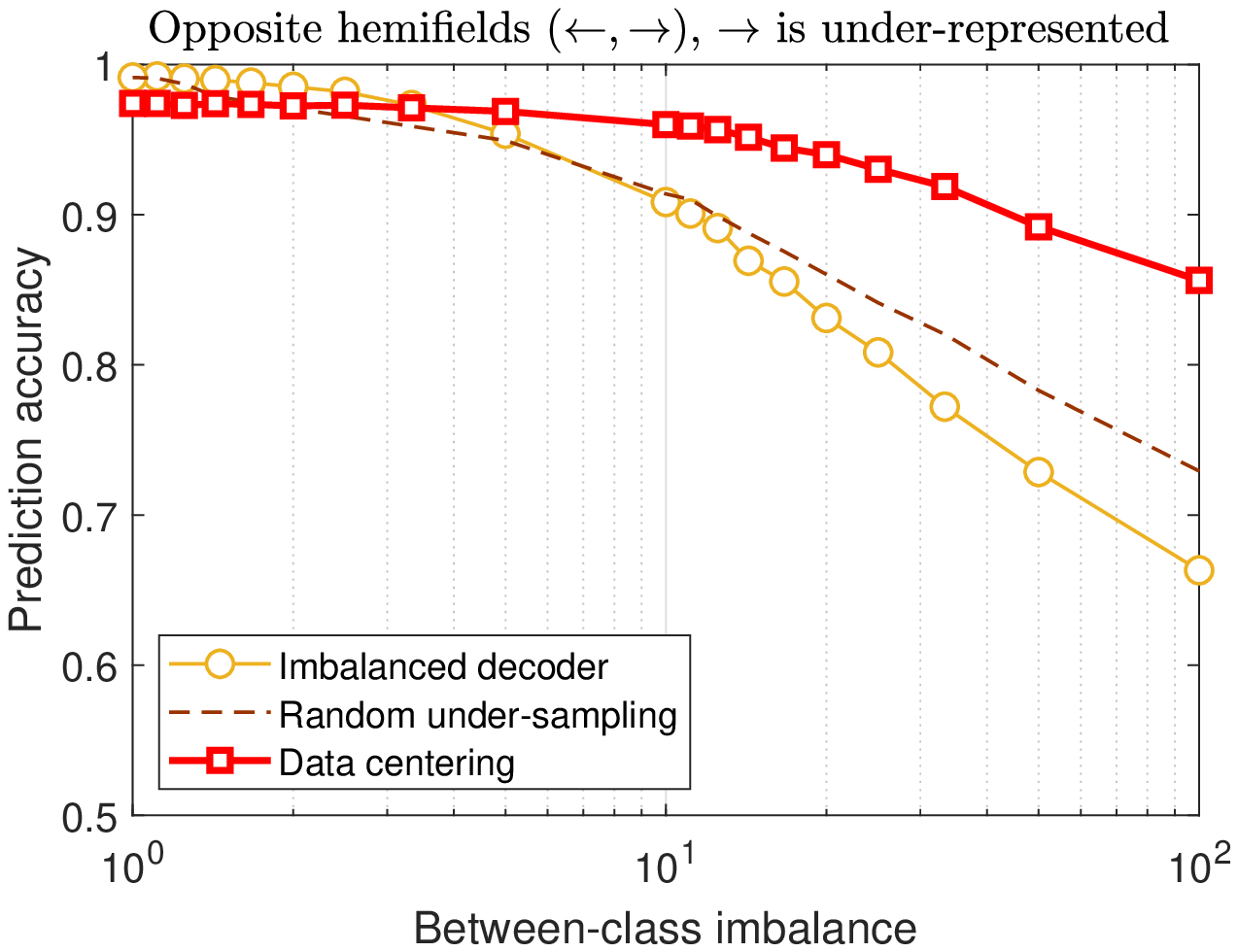}\label{results4a}}
\hfil
\subfloat[EDC-$19$ as source]{\includegraphics[scale=0.55]{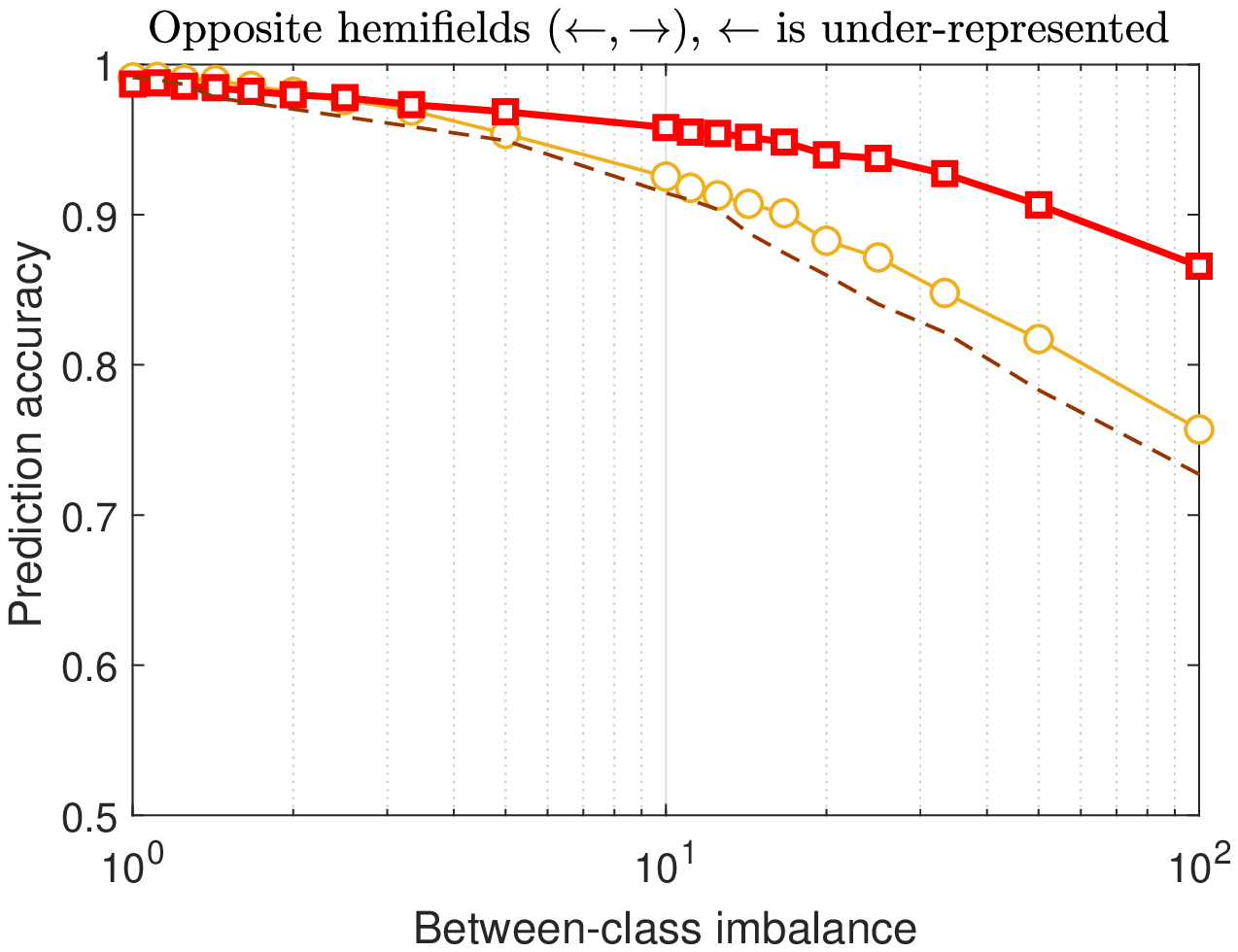}\label{results4b}}
\caption{{Average cross-subject decoding with imbalanced data sets. EDC=$6$ in Monkey A is the destination, while the source EDC is optimized and given in the captions of each plot. The eye movement directions are given in the titles of the plots where the under-represented class is highlighted. The legend in (a) applies to (b) as well.  $T = 650$, $L=2$ and $\alpha=1$.}}
\label{results4}
\end{figure}

To begin with, we consider the eye movement directions $(\leftarrow, \rightarrow)$; in this case, the decoding performance when the data is balanced is close to perfect, i.e., close to $100\%$ prediction accuracy.
The results are shown in Figs.~\ref{histograms} and \ref{results4}; we note that the optimal source EDC for the unbalanced class, i.e., the one that maximizes the decoding performance was found via exhaustive search, and is given in the caption of the plots.
The histograms in Fig.~\ref{histograms} show the distribution of prediction accuracy per target for $1000$ random subsets for the under-represented target and between class imbalance of 100; considering that the average number of trials per target (for data clustering window of $900$ and $8$ targets) is just above $100$, this imbalance ratio amounts to only few, i.e., $1$ or $2$ trials for the under-represented class. We see that if we train an imbalanced decoder, the poorly represented target will be rarely decoded correctly. In other words, the decoder will most of the time assign the test data points to the well-represented class. We then observe dramatic shift of the accuracy distribution for the under-represented class, well above random choice decoder ($0.5$ prediction accuracy), after data centering. Interestingly, we also observe improvement in the accuracy for the well represented target. We attribute this to the neurological component of the data and the existence of inherent correlations and similarities between the feature spaces of the subjects.  

\begin{figure}[t]
\centering
\subfloat[EDC-$13$ as source]{\includegraphics[scale=0.55]{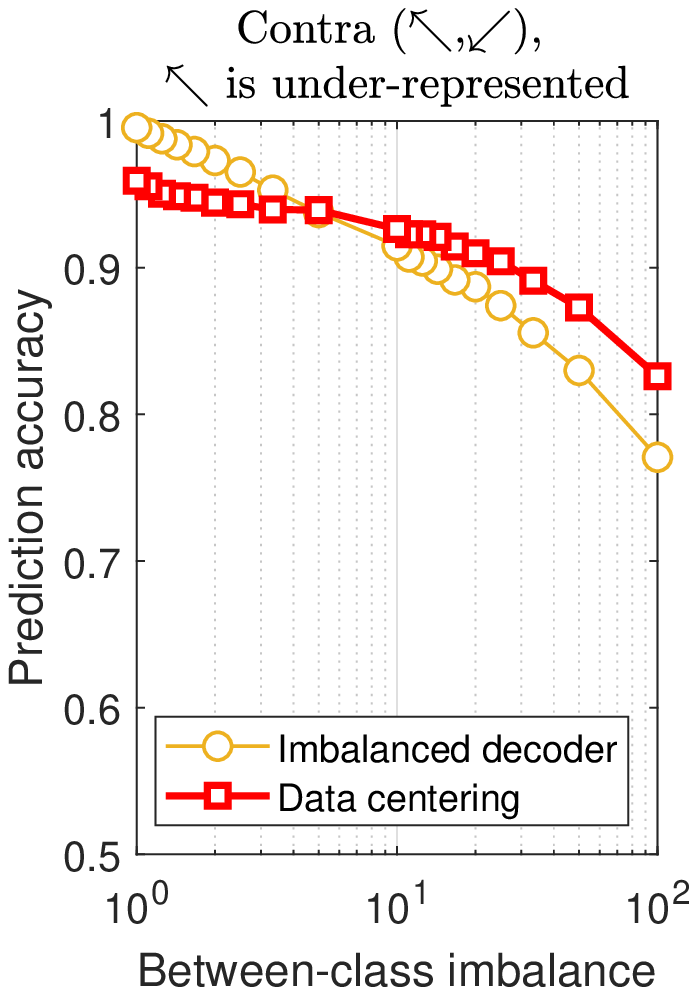}\label{results5c}}
\hfil
\subfloat[EDC-$33$ as source]{\includegraphics[scale=0.55]{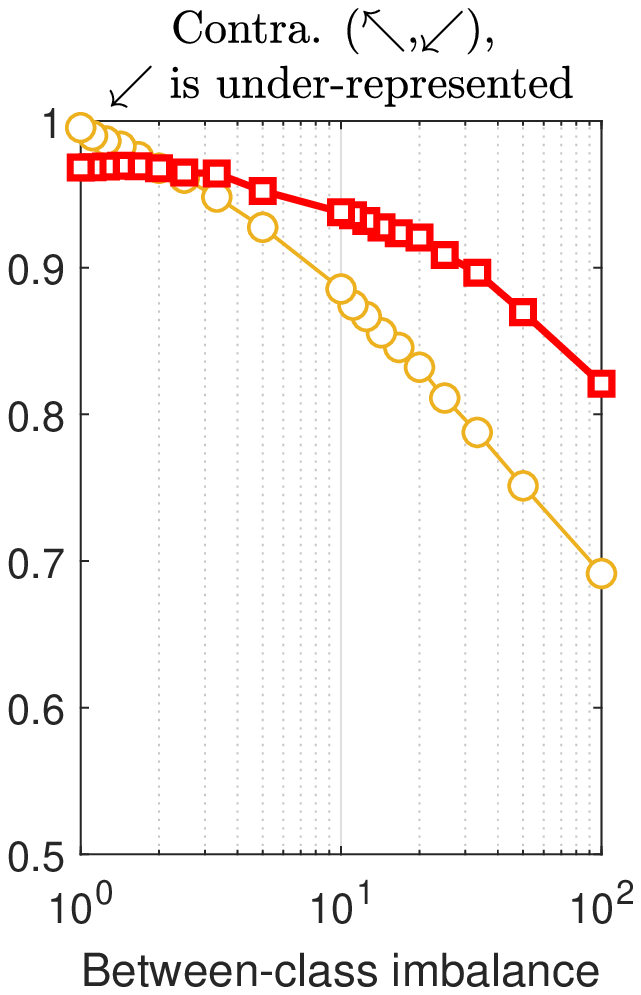}\label{results5d}}
\hfil
\subfloat[EDC-$10$ as source]{\includegraphics[scale=0.55]{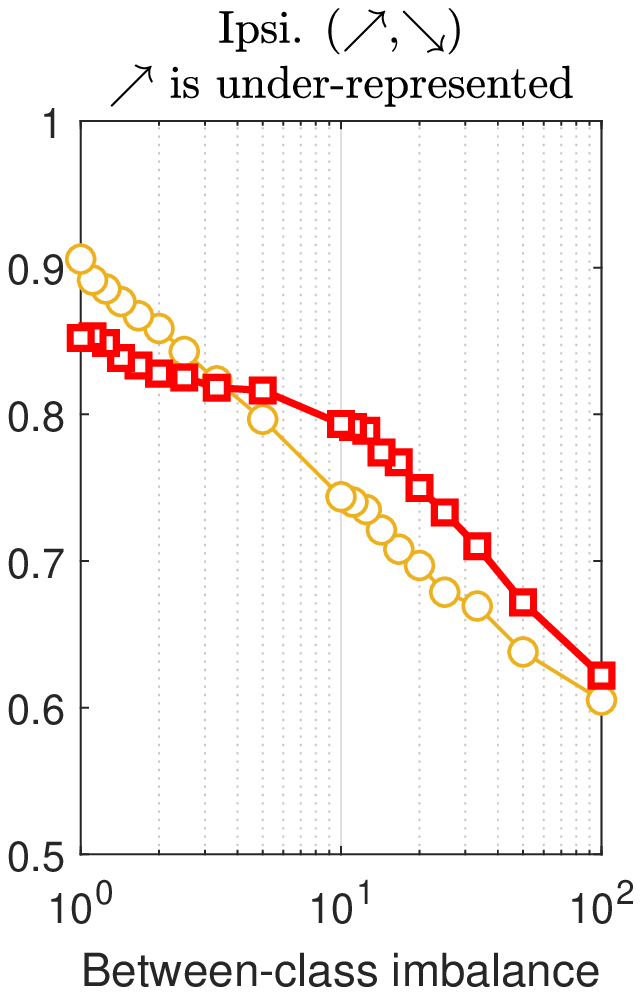}\label{results5e}}
\hfil
\subfloat[EDC-$5$ as source]{\includegraphics[scale=0.55]{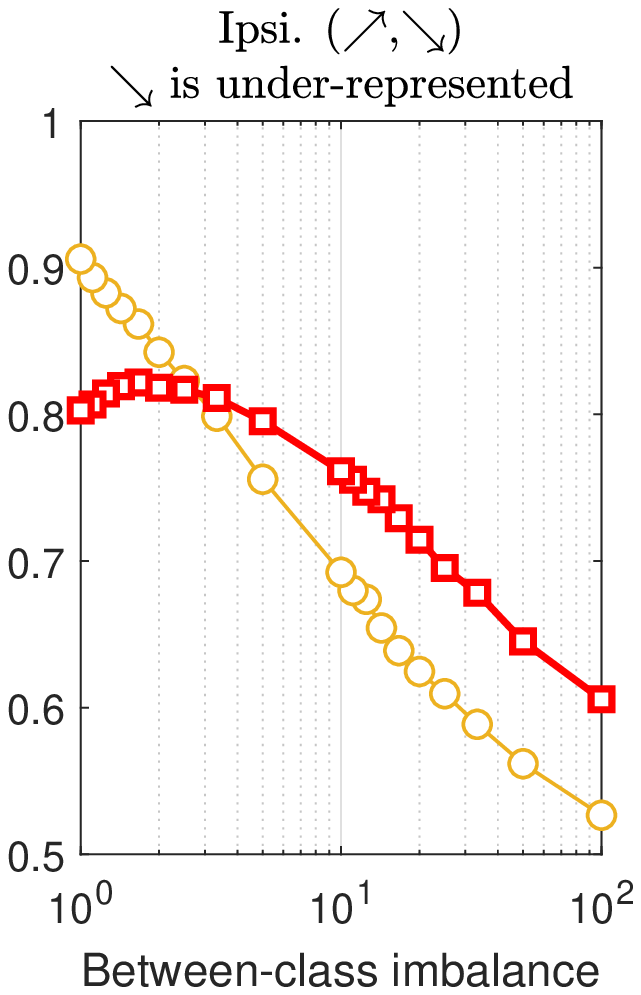}\label{results5f}}
\caption{\RI{Impact of hemifield location on average cross-subject decoding with imbalanced data sets. EDC-$6$ in Monkey A is the destination, while the source EDC is optimized and given in the captions of each plot. The eye movement directions are given in the titles of the plots where the under-represented class is highlighted. The legend in (a) applies to (b), (c) and (d) as well. $T = 650$, $L=2$ and $\alpha=1$.}}
\label{results5}
\end{figure}

Fig.~\ref{results4}, depicting only the average performance across targets, verifies the above conclusions for range of imbalance ratios; in addition, data centering also outperforms standard random sampling methods for solving the imbalanced data problem such as random undersampling of the well-represented class, as shown in Fig.~\ref{results4} as well as random oversampling via bootstrap or synthetic sampling with data generation \cite{He2009}.
We also see that as the between-class imbalance is decreased by improving the representation of the under-represented class, the gain from data centering eventually vanishes; in fact, for small between-class imbalances, local decoder training outperforms training with cross-subjected centered data for the under-represented class; this is in line with the conclusions from Figs.~\ref{results3} and~\ref{results2}.

In the setup investigated in Fig.~\ref{results4}, both movement directions belong to different hemifields. On the other hand, \cite{Angjelichinoski2019,Markowitz18412} have reported significant discrepancy in per-target decoding accuracy, depending on whether the eye movement direction was in the ipsilateral or contralateral hemifield. To assess the impact of the hemifield location of the corresponding eye movements on data centering, we consider several combinations of targets as detailed in Fig.~\ref{results5}.
By comparing the result in Figs.~\ref{results5c} and \ref{results5d} with the results in Fig.~\ref{results5e} and \ref{results5f}, we see that aforementioned discrepancy between contralateral and ipsilateral directions remains consistent for data centering.

\section{Discussion}\label{sec:conc}
\RI{We addressed the problem of cross-subject decoding from LFP data where the training data, collected in the source subject is used to decode intended motor actions by a destination subject.
As well known, neuronal activity signals are highly non-stationary over time, space and across subjects, which imposes a major challenge for cross-subject decoding; namely, direct application of a decoder trained over data collected from the source subject to decode intended actions by the destination subject, without accounting for the non-stationary nature of the data, will yield poor decoding performance.
To address the issue of non-stationary LFP data across subjects and foster reliable cross-subject decoding, we proposed novel transfer learning method, which we named data centering. The method is used on subjects that perform the exact same set of motor tasks and operates by adapting the source feature space to the destination feature space in a supervised manner. The key ingredient of data centering are the transfer functions which model the functional transformations between target-conditional distributions corresponding to the same motor intentions performed by two different subjects under the same conditions. In addition, we also developed a simple, yet effective estimation technique for linear transfer functions based on the first and second order moments of the target-conditional distributions.
In comparison with other domain adaptation transfer learning methods, data centering with linear transfer functions is simple and computationally efficient. 
}

\RI{We tested our data centering method with linear transfer functions on data collected from two macaque monkeys (Monkey A and Monkey S) performing memory-guided visual saccades to one of eight target locations. In our implementation, we applied data centering on feature spaces generated using Pinsker's method which has been recently shown to perform exceptionally well for extracting features from noisy LFP data. We first investigated the setup where each target-conditional distribution is equally represented in the feature space, i.e., the data sets are balanced in both the source and the subjects. The results have shown a peak cross-subject decoding performance of $80\%$ and $60\%$ in Monkey A and Monkey S, respectively, even with a simple linear classifier such as LDA; the peak performance was achieved at superficial cortical depths in the destination subject. This marks a substantial improvement over random choice decoder, proving that data-centering is essential for the success of LFP-based cross-subject decoders. In addition, we concluded that (i) for balanced data sets, the cross-subject decoding performance is dominated by, and remains consistently below the subject-specific performance of the destination, (ii) the geometry of the destination feature space strongly impacts the cross-subject decoding performance, and (iii) the dependence of cross-subject decoding performance on the source data set is loose.
}

\RI{Our second study focused on the case of imbalanced data sets, i.e., the setup where one or more target-conditional distributions are under-represented. Specifically, we investigated whether data centering can be used for correcting the balance between the target-conditional distributions by bringing data for the under-represented class(es) from another, source subject, which performs the same set of motor actions. We conducted our experiments in a simplified, binary setup where the subjects perform visual saccades to one of two possible target locations and we optimized the decoding performance over the source data set via exhaustive search. The results show that data centering indeed improves the decoding performance in imbalanced data sets; the improvement is particularly dramatic for eye movement directions in opposite hemifields and large between-class imbalance ratios (going up to $100$, which corresponds to $1-2$ trials for the under-represented class), where we observed relative improvement gains as high as $40\%$ over the decoder trained using the imbalanced data set. In addition, we have also empirically shown that data centering outperforms standard sampling methods for dealing with imbalanced data sets.
These results indicate the potential application of data centering in neural prosthesis where the objective is the restoration of lost motor function in individuals with chronic disabilities, confirming the promising practical potential of data centering.}


\ifCLASSOPTIONcompsoc
  \section*{Acknowledgments}
\else
  \section*{Acknowledgment}
\fi

This work was supported by the Army Research Office MURI Contract Number W911NF-16-1-0368. 

\bibliographystyle{IEEEtran}
\bibliography{bare_jrnl}



%


\end{document}